\documentclass[sigconf, screen]{acmart}

\makeatletter                   
\def\mdseries@tt{m}             
\makeatother                    
\usepackage[plain]{fancyref}
\usepackage[draft=true]{minted} 
\usepackage{color}
\usepackage{hyperref}           
\hypersetup{
    colorlinks=true,
    linkcolor=blue,
    filecolor=red,      
    urlcolor=magenta,
    breaklinks=true,            
}
\usepackage{breakurl}           

\renewcommand\footnotetextcopyrightpermission[1]{} 
\acmConference[]{}{}{} 
\acmYear{}
\acmDOI{}
\acmISBN{}
\acmPrice{} 

\usepackage{graphicx}       
\usepackage{comment}        
\usepackage{amsmath}        
\usepackage{hyperref}       
\usepackage{subfloat}       
\usepackage{subcaption}
\usepackage{multirow}

\usepackage{algorithm}
\usepackage{algpseudocode}
\usepackage{amsmath}

\usepackage{graphicx}
\DeclareGraphicsExtensions{.pdf,.png,.jpg}

\usepackage{subcaption}
\usepackage{tikz}
\usepackage{adjustbox}

\usepackage{todonotes}      

\newif\ifcomm
\commtrue
\ifcomm
\else
\commfalse
\fi
\ifcomm
	\newcommand{\mycomm}[3]{{\footnotesize{{\color{#2} \textbf{[#1: #3]}}}}}
	\newcommand{\CRdel}[1]{\textcolor{red}{\sout{#1}}}
\else
    \newcommand{\mycomm}[3]{}
    \newcommand{\CRdel}[1]{}
\fi
\newcommand{\rana}[1]{\mycomm{Rana}{purple}{#1}}

\newtheorem*{theorem*}{Theorem}

\newcommand{\alg}{\emph{LAMPS}}
\newcommand{\algFULL}{\emph{LLM API- and Memory-based Predictive Scheduling}}

\begin{document}
\sloppy 

\title{Fast Inference for Augmented Large Language Models}

\date{}

\author{Rana Shahout$^{\dagger}$, Cong Liang$^{\mathsection}$, Shiji Xin$^{\dagger}$, Qianru Lao$^{\dagger}$, \\Yong Cui $^{\mathsection}$, Minlan Yu$^{\dagger}$, Michael Mitzenmacher$^{\dagger}$}

\affiliation{
  \institution{Harvard University~$^{\dagger}$, Tsinghua University~$^{\mathsection}$}
}

\begin{abstract}
Augmented Large Language Models (LLMs) enhance the capabilities of standalone LLMs by integrating external data sources through API calls. In interactive LLM applications, efficient scheduling is crucial for maintaining low request completion times, directly impacting user engagement. However, these augmentations introduce scheduling challenges due to the need to manage limited memory for cached information (KV caches). As a result, traditional size-based scheduling algorithms, such as Shortest Job First (SJF), become less effective at minimizing completion times.
Existing work focuses only on handling requests during API calls by preserving, discarding, or swapping memory without considering how to schedule requests with API calls.
In this paper, we propose \alg{}, a novel LLM inference framework for augmented LLMs. \alg{} minimizes request completion time through a unified scheduling approach that considers the total length of requests and their handling strategies during API calls. Recognizing that LLM inference is memory-bound, our approach ranks requests based on their consumption of memory over time, which depends on both the output sizes and how a request is managed during its API calls. To implement our scheduling, \alg{} predicts the strategy
that minimizes memory waste of a request during its API calls, aligning with but improving upon existing approaches. We also propose starvation prevention techniques and optimizations to mitigate the overhead of our scheduling.
We implement \alg{} on top of vLLM and evaluate its performance against baseline LLM inference systems, demonstrating improvements in end-to-end latency by 27\%-85\% and reductions in TTFT by 4\%-96\% compared to the existing augmented-LLM system, with even greater gains over vLLM.

\end{abstract}

\maketitle
\section{Introduction}

Recent progress in large language models (LLMs) has initiated a new wave of interactive AI applications. A prominent example is OpenAI's ChatGPT~\cite{openai2022chatgpt}, which facilitates conversational interactions across various tasks. One way to extend LLM capabilities is to augment them with external tools~\cite{mialon2023augmented}, resulting in what we refer to as API-augmented requests. These augmentations include arithmetic calculation~\cite{hao2024toolkengpt}, ChatGPT plugins~\cite{ChatGPTplugins}, image generation~\cite{betker2023improving}, and virtual environments~\cite{shridhar2020alfworld}. Consequently, AI development is increasingly moving towards compound AI systems~\cite{compound-ai-blog} that integrate multiple interacting components, such as model calls, retrievers, and external tools, rather than relying solely on monolithic models.

API-augmented requests present several challenges, particularly regarding memory consumption during the LLM decoding phase, which is memory-bound and requires careful management. Each request has associated key and value matrices that grow in size during the request. LLMs cache these matrices in a key-value (KV) cache throughout the sequence generation process to enhance efficiency, eliminating the need to recompute them at every iteration. This caching significantly reduces computation time but requires substantial memory, roughly proportional to the input and output lengths, the number of layers, and hidden dimensions. High memory consumption during decoding can translate to higher latency and lower throughput, as it limits the system's ability to process multiple requests concurrently.

\begin{figure}[t]
    \centering
    \begin{subfigure}[b]{0.32\linewidth}
        \centering
        \includegraphics[width=\textwidth]{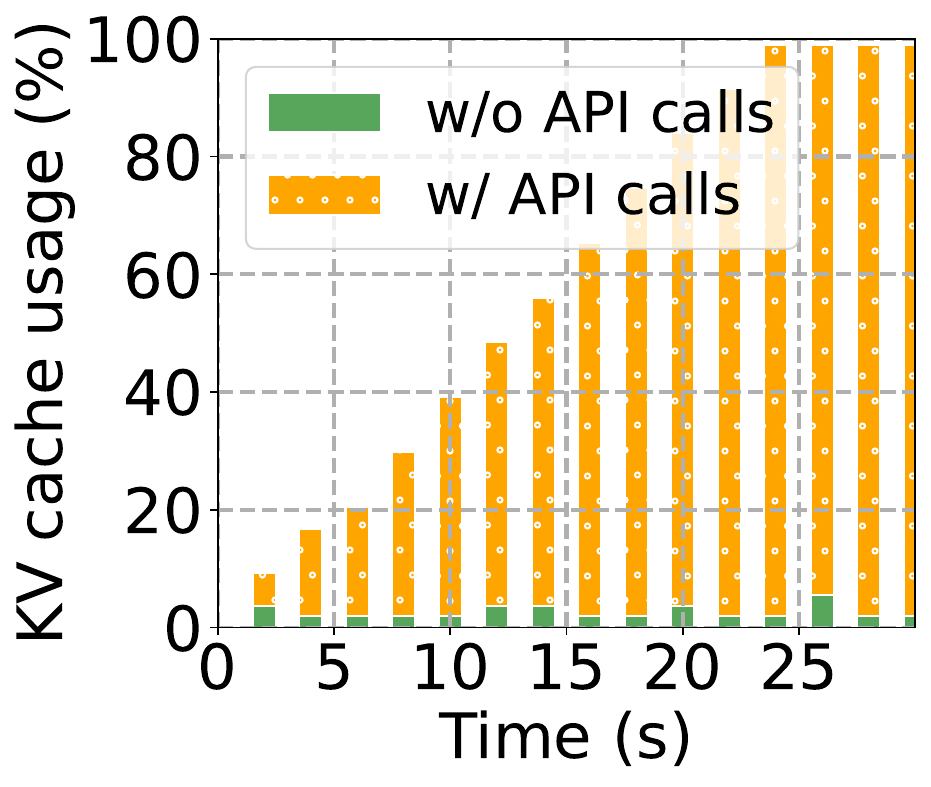}  
        \caption{Preserve}
        \label{}
    \end{subfigure}
    \hfill
    \begin{subfigure}[b]{0.32\linewidth}
        \centering
        \includegraphics[width=\textwidth]{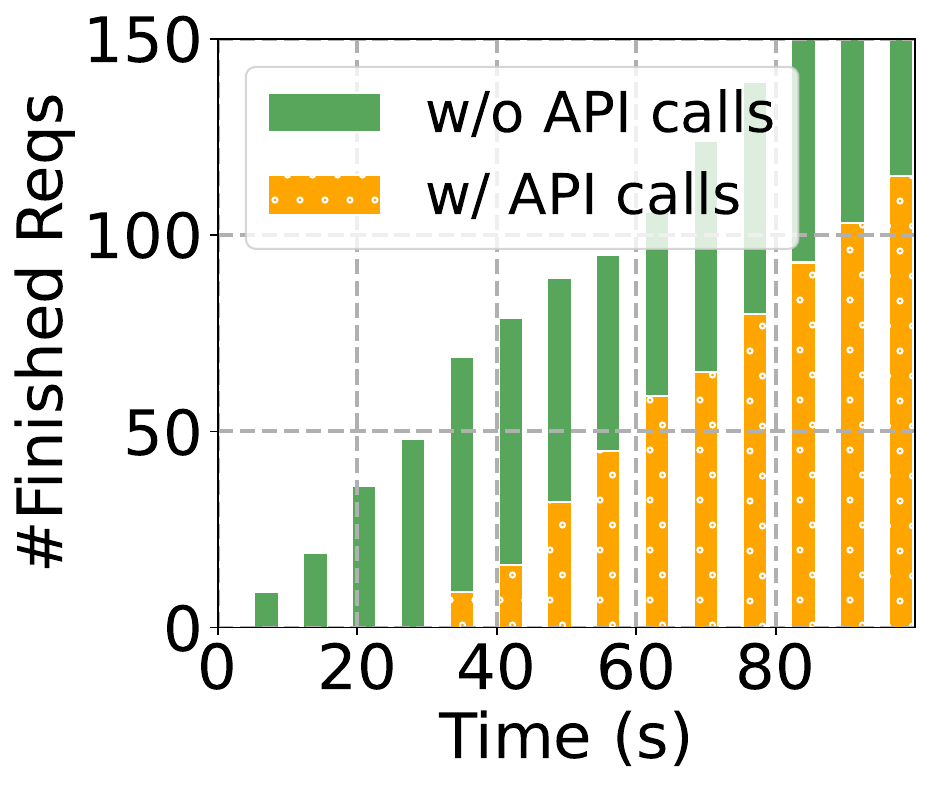}  
        \caption{Preserve}
        \label{}
    \end{subfigure}
    \hfill
    \begin{subfigure}[b]{0.32\linewidth}
        \centering
        \includegraphics[width=\textwidth]{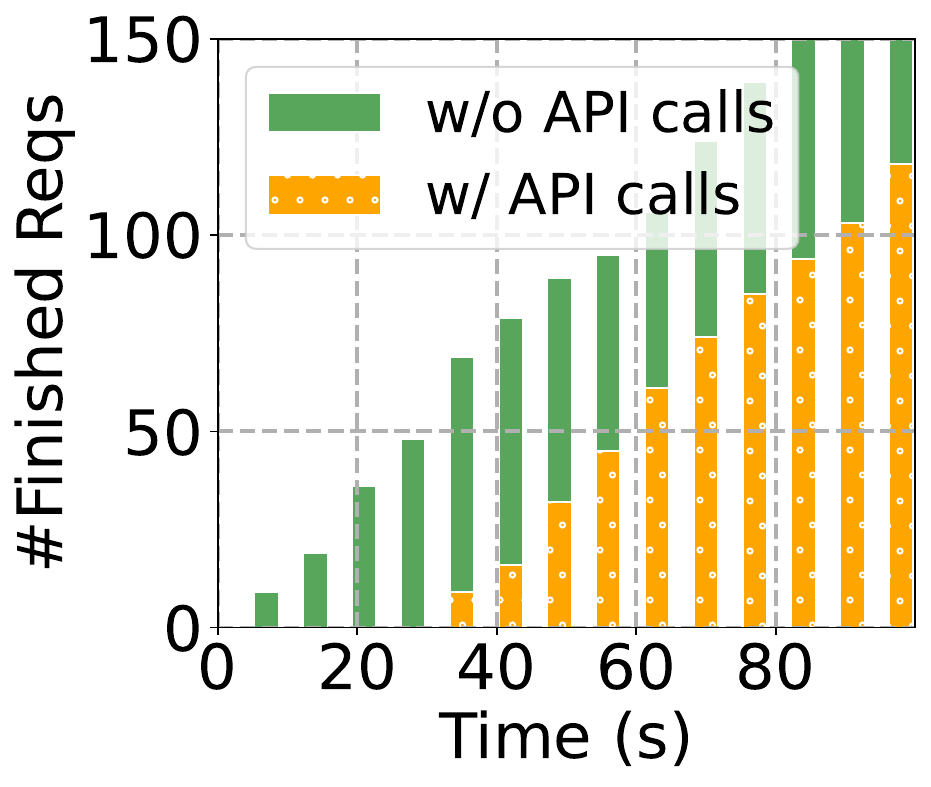}  
        \caption{Discard}
        \label{}
    \end{subfigure}
    \caption{Impact of including API calls: using a subset of INFERCEPT~\cite{abhyankarinfercept}, we compare two variations of the dataset—one with API calls and one without. (a) KV cache usage (\%) over time when all API calls are handled using Preserve. (b) Number of completed requests over time using Preserve. (c) Number of completed requests over time using Discard.}
    \label{fig:bad_KV}
\end{figure}

With API augmentation, memory demands increase further, depending on how the system handles requests during API calls (Figure~\ref{fig:bad_KV}).
There are three main strategies for handling a request's KV cache during API calls:
\begin{itemize}
    \item 
\emph{Preserve}: The system retains the KV cache in memory while waiting for the API response.
    \item 
\emph{Discard and Recompute}: The system discards and recomputes the KV matrices from the start once the API returns.
    \item 
\emph{Swap}: The system offloads the KV cache to the CPU to free up memory, and reloads it when the API returns.
\end{itemize}

We refer to these as {\em memory handling strategies}, or {\em handling strategies} for brevity. All three strategies have downsides. With \emph{Preserve}, because the KV cache remains in memory throughout the API call,
memory is wastefully consumed during the call. \emph{Discard and Recompute} incurs additional memory and computational costs when recomputing the KV cache. \emph{Swap} introduces overhead from pausing running requests and managing data transfer between CPU and GPU memory. Additionally, because the duration of API calls can vary significantly across different augmentation types and requests, one strategy does not fit all requests. For instance, a simple calculation might be completed in milliseconds, whereas image generation could take several seconds.

Existing LLM inference systems generally operate for standalone LLMs and fail to guarantee low request completion time for augmented LLMs. This issue becomes more significant when handling requests involving external API calls, leading to delays such as head-of-line (HoL) blocking. HoL blocking occurs when long-running requests, including those waiting for API responses, prevent shorter ones from being processed efficiently.
Previous works attempt to mitigate the issue of HoL blocking by managing memory during API calls.
For example, vLLM~\cite{kwon2023efficient} discards and recomputes API-augmented requests, treating API calls as termination signals and the request returning from the API as a new job. INFERCEPT~\cite{abhyankarinfercept} classifies requests and applies strategies dynamically, aiming to reduce memory wastage for API-augmented requests. However, both systems still rely on a first-come first-served (FCFS) scheduling policy, which increases HoL blocking by allowing long-running requests to block shorter ones, especially under high load.

Scheduling strategies can reduce head-of-line blocking by prioritizing requests. Traditional size-based scheduling prioritizes jobs\footnote{The terms job and request are used interchangeably in this paper.}  with shorter execution time.  Without API calls, this approach is effective, as execution time and memory usage for LLM requests without API calls correlate (with the execution time corresponding roughly to the output length)~\cite{trail}. However, with API-augmented requests, the output length may not reflect the total request time, including API calls.
A request with a short output might involve a lengthy API call, while a request with a longer output may require minimal API interaction. 

Rather than treating LLM execution and API calls as separate processes, we argue that integrating the scheduling and memory handling of requests can significantly improve request completion time, especially under high load. Achieving this requires information about both request output length and API call time, which are often unavailable, so we rely on predicting these values. This challenge leads to a key question: \emph{Given predictions of request output length and API call time, how should we schedule and handle API-augmented requests to prevent head-of-line blocking and minimize latency?}

This paper presents \alg{} (\algFULL{}), a novel inference framework for augmented LLMs.
\alg{} minimizes request completion time through a unified scheduling approach that jointly considers the total length of requests and their API call-handling strategies.
Recognizing that LLM inference is memory-bound, our approach ranks requests based on their \emph{memory over time}, which reflects how memory resources are allocated throughout a request’s processing. By considering both the output size and the memory handling strategies during API calls, \alg{} enables effective management of varying output sizes and API interactions.

To achieve this, \alg{} predicts memory consumption using request properties, such as pre-API output length and API characteristics. We develop a prediction model leveraging the opt-125m language model~\cite{zhang2022opt} to estimate pre-API output lengths from input prompts while predicting API duration based on the type of API being called.
Based on this profile, we determine which handling strategy to use for a request: Preserve, Discard and Recompute, or Swap.
If a request involves multiple API calls, each call is treated independently, and the request is reinserted into the scheduling process after each API call is handled, continuing until the last API call is completed.
Our approach aligns with INFERCEPT equations~\cite{abhyankarinfercept}, except we determine requests' memory strategy {\em before} scheduling them. We then schedule the requests according to our policy, which aims to minimize response time by prioritizing requests by their memory footprint over time, accounting for API interactions.
While it is theoretically possible to collectively optimize handling and scheduling decisions, such an approach is impractical in an online setting. Consequently, we employ a greedy algorithm that first minimizes memory usage for each individual request and then schedules requests based on their memory requirements over time.

Our contributions are as follows: 
\begin{itemize}
    \item We develop a prediction model to estimate pre-API output lengths from input prompts and predict API duration based on the type of API being called, allowing us to predict memory consumption over time.
    \item Using the predicted memory consumption over time, we assign a handling strategy to each request before processing.
    \item We propose a scheduling policy that considers the request length and the API handling strategy to minimize request completion time. We integrate optimizations into our scheduling policy to mitigate starvation and reduce scheduling overhead.
    \item We implement \alg{} on top of vLLM~\cite{kwon2023efficient}, a state-of-the-art LLM inference system. We evaluate \alg{} on two datasets, comparing \alg{} against baseline systems.
    Our results show that \alg{} consistently outperforms INFERCEPT across various datasets and request rates, achieving improvements in end-to-end latency ranging from 27\% to 85\% and reductions in TTFT from 4\% to 96\%, with even greater improvements over vLLM.
    We also analyze the components of \alg{} and the effect of prediction accuracy on its performance.
\end{itemize}

\section{Background and Motivation}


LLMs expand their capabilities by integrating external tools, allowing them to handle more complex tasks. However, this augmentation introduces challenges for request handling and scheduling with the goals of minimizing response times and managing memory efficiently during inference. This section presents the background for LLM \mbox{execution and API interactions.}

\subsection{Augmented LLMs}
\begin{figure}[t]
  \centering
  \includegraphics[width=\linewidth]{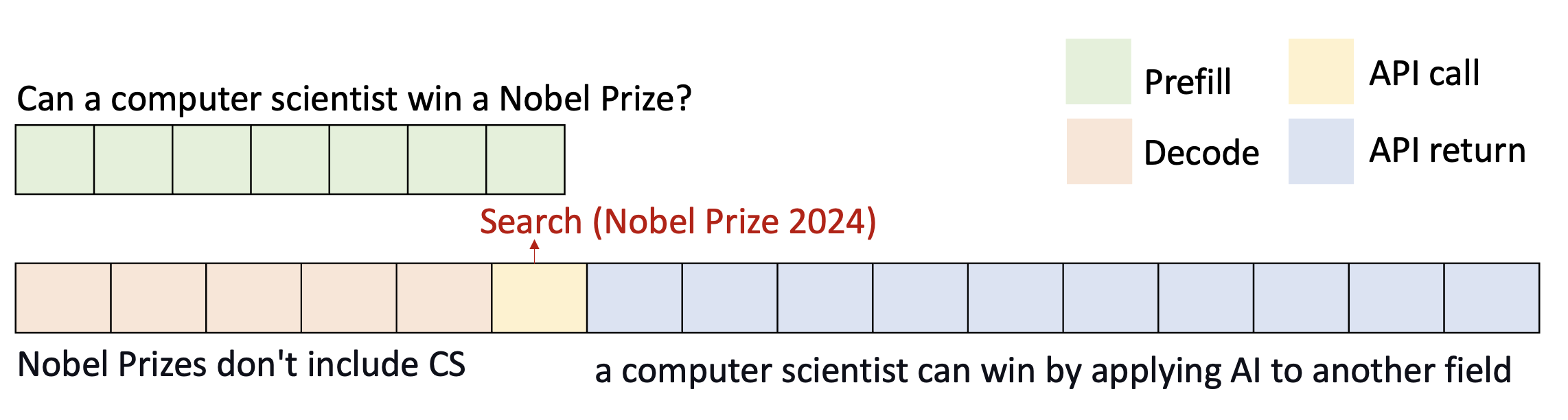}
  \caption{Illustration of an augmented-LLM request. The API fetches detailed information about the 2024 Nobel Prize.}
  \label{fig:augmented_llm}
\end{figure}

Augmented Language Models~\cite{mialon2023augmented, wang2024survey} refer to language models that enhance their capabilities by incorporating external tools, retrieval mechanisms, or reasoning strategies to overcome the limitations of traditional LLMs. Unlike pure LLMs, which rely solely on pre-trained parameters to generate responses, augmented LLMs can query external data sources to expand their capabilities.
Figure~\ref{fig:augmented_llm} shows an example of an augmented LLM request. These augmentations, which we refer to as \emph{API} (Application Programming Interfaces), fall into three main categories as described in~\cite{mialon2023augmented}: incorporating non-LLM tools during decoding (such as calculators~\cite{wolfarm}, information retrieval systems~\cite{baeza1999modern}), iterative self-calling of an LLM (like chatbots maintaining conversation history), and complex compositions involving multiple LLMs, models, and tools (exemplified by frameworks like LangChain~\cite{langchain}, DSpy~\cite{khattab2024dspy}, Gorilla~\cite{patil2023gorilla}, SGLang~\cite{zheng2023efficiently}, and AgentGraph~\cite{chen2019agentgraph}).

LLM API time varies significantly based on augmentation types, with a clear distinction between short-running and long-running augmentations.
Despite this variation, today's systems still rely on FCFS scheduling.
This suggests that API handling strategies should be tailored to specific augmentation types rather than using a one-size-fits-all approach.

\subsection{Transformer-Based Generative Models}

At each step, a Transformer model generates the most probable next token based on the sequence of previously generated tokens. A model generating a sequence of length $n$ needs to perform $n$ iterations, with each token passing through several layers of self-attention and feed-forward networks.

During the $i$-th iteration, the model operates on all prior tokens ($t_0, t_1, \dots, t_{i-1}$) using self-attention mechanisms. The resulting output can be represented as:

\[
h_{\text{out}} = \text{softmax}\left(\frac{q_i \cdot K^\top}{\sqrt{d_h}}\right) \cdot V
\]

Here, $q_i$ is the query vector for the current token $t_i$, while $K$ and $V$ are matrices containing the key and value vectors for all preceding tokens, where $K, V \in \mathbb{R}^{i \times d_h}$.

\subsubsection{Key-Value (KV) Cache} 
To reduce computational overhead, LLMs cache the key and value matrices (KV cache) during sequence generation. This approach avoids recomputing these matrices at each step, improving efficiency but leading to high memory usage, which scales with the sequence length, number of layers, and hidden dimensions. As more tokens are generated, memory demands grow, particularly for long sequences. For instance, the GPT-3 175B model requires around 2.3 GB of memory to store key-value pairs for a sequence length of 512 tokens. This high memory consumption poses challenges for efficient preemptive scheduling, especially when working with limited GPU memory.

\subsubsection{Scheduling in LLMs} 

In LLM inference systems, iteration-level scheduling, as implemented in systems like Orca~\cite{yu2022orca} and vLLM~\cite{kwon2023efficient}, is commonly used. 
It differs from traditional request-level scheduling, where the system processes a batch of requests until completion, forcing earlier requests to wait until the entire batch is completed, and new requests must wait in a queue until the next batch is processed. Iteration-level scheduling processes one token at a time for each request in the batch, allowing the system to dynamically adjust the batch after every iteration. Requests that complete an iteration can exit the batch, and new ones can be introduced, optimizing resource usage within the constraints of GPU memory. The default policy in these systems is FCFS.
Most recent scheduling works~\cite{trail, wu2023fast}, however, focus on LLM requests without API augmentations. API-augmented requests introduce challenges, such as handling external interactions and variable API call times, which require new scheduling strategies beyond those used for standalone LLMs. Accordingly, \alg{} focuses on developing such scheduling for  API-augmented requests.

\subsection{Handling Requests During API}
\label{sec:background_api_overhead}

Optimizing memory management during API calls involves selecting a strategy that minimizes GPU memory waste. In an augmented LLM inference system, this choice depends on two factors: the duration of the API call and the length of the pre-API output.
For brief API calls, the Preserve strategy may be advantageous in avoiding recomputation overhead. For longer API calls, either Discard or Swap is preferable. If the pre-API portion of the request is computationally light (short), Discard is beneficial. Otherwise, Swap may be more efficient despite the potential delays it introduces.

INFERCEPT addresses this optimization challenge by developing the following equations (equations 1-3 in~\cite{abhyankarinfercept}) that model the memory wastage associated with each handling strategy. 
\begin{align}
\text{WastePreserve}_{i} &= T_{\text{INT}} \times C_{i} \times M \label{eq:waste_preserve}
\end{align}
\begin{align}
\text{WasteDiscard}_{i} &= T_{\text{fwd}}(C_{i}) \times C_{i} \times M + T_{\text{fwd}}(C_{i}) \times C_{\text{other}} \times M \label{eq:waste_discard}
\end{align}
\begin{align}
\text{WasteSwap}_{i} &= 2 \times T_{\text{swap}}(C_{i}) \times C_{\text{batch}} \times M \label{eq:waste_swap}
\end{align}
Here $T_{\text{INT}}^{j}$ is the duration of the API call for request $i$, and $C_{i}$, $C_{\text{other}}$, and $C_{\text{batch}}$ represent the context size (in tokens) of request $i$ before the API call, the context size of other requests in the batch with request $i$, and the total context size of all requests in the batch, respectively. $M$ denotes the memory consumed per token for the KV cache. $T_{\text{fwd}}(C_{i})$ and $T_{\text{swap}}(C_{i})$ represent the time required for model forwarding with context $C_{i}$ and the time to swap context $C_{i}$, respectively. INFERCEPT dynamically selects a strategy that minimizes memory waste. However, its scheduling policy remains FCFS.

\section{Challenges and Design Principles}


\begin{table}[t]
  \centering
  \begin{tabular}{|c|c|c|c|}
    \hline
     & $R_1$ & $R_2$ & $R_3$ \\ \hline
    Total length   & 6   & 2   & 3   \\ \hline
    API start after   & 5  & 1  & 2  \\ \hline
    API duration   & 2  & 7  & 1  \\ \hline
    Memory action   & Preserve  & Discard  & Swap  \\ \hline
  \end{tabular}
  \caption{Total length and API call duration in iterations unit.}
  \label{tab:example_durations}
\end{table}

\begin{figure}[t]
  \centering
  \begin{tabular}{cc}
    \subfloat[FCFS]{\includegraphics[width=\columnwidth]{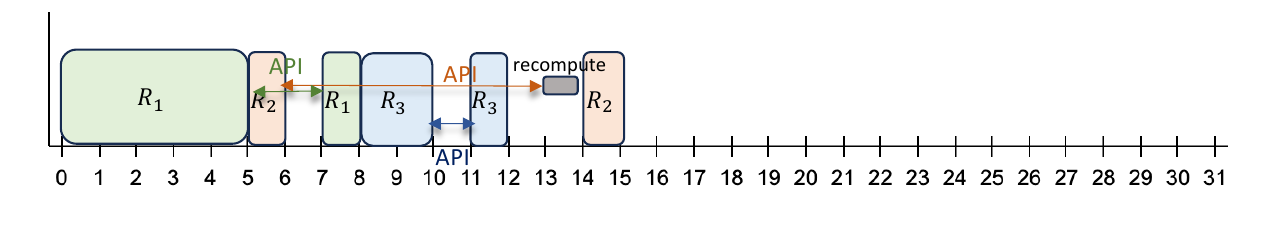} \label{fig:example_fcfs}}
 \\
    \subfloat[SJF (request length)]{\includegraphics[width=\columnwidth]{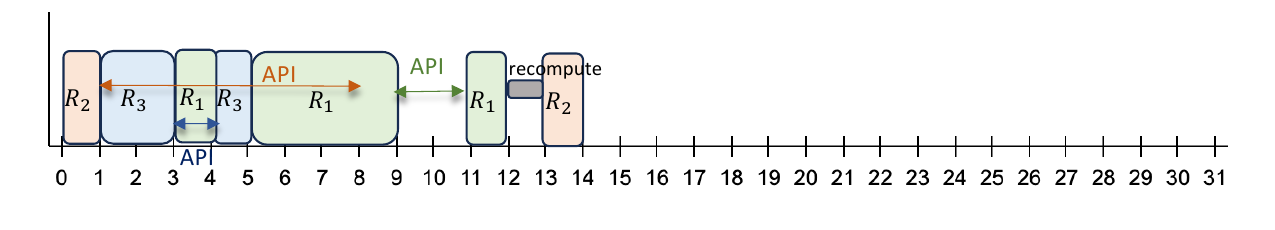} \label{fig:example_sjf}}  \\
    \subfloat[SJF (request+API length)]{\includegraphics[width=\columnwidth]{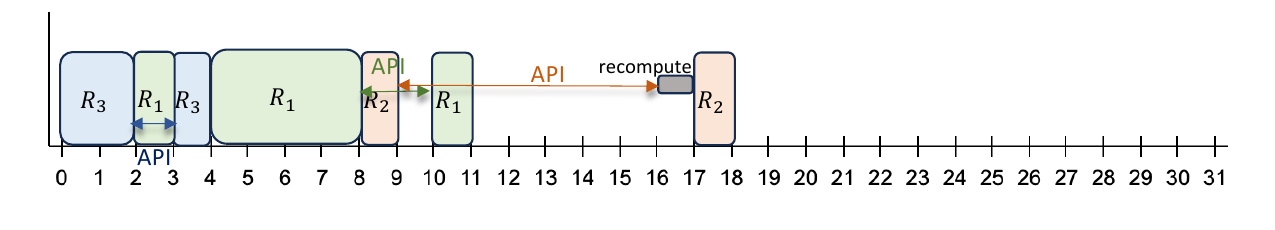} \label{fig:example_stl}} \\
    \subfloat[Preferred*]{\includegraphics[width=\columnwidth]{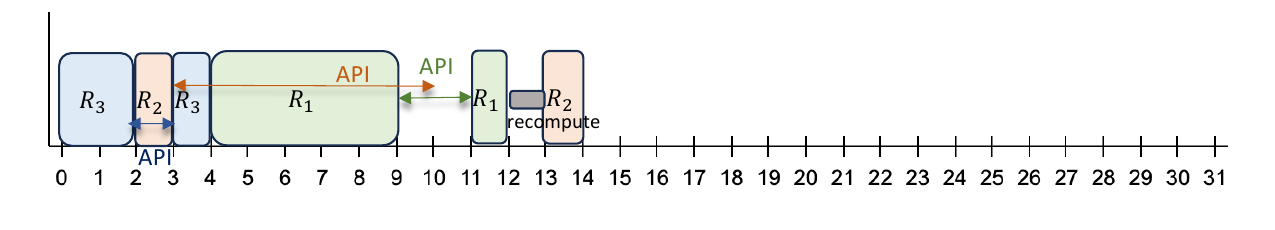} \label{fig:example_opt}} 
  \end{tabular}
  \caption{Comparison of scheduling policies for API-augmented requests with a memory budget of $6$ unit, requests lengths and API duration are summarized in Table~\ref{tab:example_durations}. (a) FCFS: yielding an average request completion time of 11.66 units. (b) SJF: achieving an average request completion time of 10.33 units. (c) SJF by Total Length: Orders requests by total length, achieving an average request completion time of 11 units (d) Optimized: Integrates length and API handling, achieving an average request completion time of 10 units.}
  \label{fig:example}
\end{figure}

\subsection{Challenge: Scheduling API-augmented Requests}


Our goal is to reduce average and typical request times through scheduling.  Size-based scheduling methods, such as Shortest Job First (SJF), can reduce request completion time by utilizing known or predicted request sizes. Traditional scheduling methods encounter challenges with API-augmented requests. Indeed, it is not clear what the size should be with API calls: should the API delay be included are not? Even assuming known output lengths, SJF can fail to perform optimally when requests involve API calls. The following example in Figure~\ref{fig:example} illustrates this issue.



\textbf{Example.} Consider three requests R1, R2, and R3 that all arrive at time 0. Each request includes one API call, triggered at different times during decode generation. Their output lengths are 6, 2, and 3 tokens, respectively, with API durations (in token generation units) of 2, 7, and 1 units, respectively. 
The strategies to handle requests during the API for each request (determined by the INFERCEPT equations) are Preserve for R1, Discard for R2, and Swap for R3.
Table~\ref{tab:example_durations} summarizes this information.

In this example, we assume that only one request can run at a time and the memory budget is limited to $6$ units. We first consider 
the setting of INFERCEPT. The strategy for handling each request during its API call is determined dynamically at runtime using the INFERCEPT equations (~\eqref{eq:waste_preserve},~\eqref{eq:waste_discard} and~\eqref{eq:waste_swap}). For simplicity, we assume complete information is known about the total length and API duration of each request. A request is scheduled only if there is enough memory available.


Although all requests arrive at the same time, the FCFS scheduling policy used by INFERCEPT determines their order based on request ID, processing them as $R_1, R_2, R_3$.
With a memory budget of 6 units, the scheduler processes the pre-API part of $R_2$ during $R_1$'s API call, as $R_2$ will be discarded after one unit, freeing memory to continue with $R_1$. In contrast, $R_3$'s pre-API part cannot run during $R_1$'s API call because it will not release memory before the API response completes, preventing $R_1$ from resuming. This scheduling yields an average request completion time of 11.66 units (Figure~\ref{fig:example_fcfs}).
The SJF policy schedules requests based only on length, processing them from shortest to longest: $R_2, R_3, R_1$. At time unit 8, the API of $R_2$ completes, leaving a post-API part of length 2 (including recomputation). However, the running request, $R_1$, also has two units remaining, so $R_2$ must wait. At time unit 9, $R_1$ enters its API call, consuming five units of memory, leaving only 1 unit available, which is insufficient to start the post-API part of $R_2$. As a result, $R_2$ must wait until $R_1$ finishes. This policy results in an average request completion time of 10.33 units (Figure~\ref{fig:example_sjf}).

These approaches fall short of optimal scheduling because they ignore the interaction between scheduling and request handling during API calls. A naive strategy, referred to as \emph{SJF by total length} (Figure~\ref{fig:example_stl}), orders requests by their total length (output length plus API duration). Again, in this example, the pre-API part of $R_2$ can run during $R_1$'s API call. This policy achieves an average request completion time of 11 units, worse than SJF.
A more effective scheduling policy (Figure~\ref{fig:example_opt}) integrates total length and the API handling strategy. This approach yields an average request completion time of 10 units, outperforming previous methods. Notably, the post-API part of $R_2$ becomes ready at time unit 10, but due to memory constraints, it waits until $R_1$ finishes.

Our intuition is that, under memory constraints, $R_3$, the least memory-intensive request, should run first to release memory quickly. It should be followed by $R_2$, with $R_1$ (the most memory-consuming request due to its Preserve handling) scheduled last. This insight informs our proposal to incorporate API handling strategies into the scheduler, ranking requests based on memory consumption over time.


\subsection{Key Design Principles}
\label{sec:opportunities}

\begin{table}[t]
    \centering
    \small
    \begin{tabular}{|c|l|c|c|}
        \hline
        \textbf{Dataset} & \textbf{Type} & \textbf{Duration (sec)} & \textbf{Num} \\ \hline
        \multirow{6}{*}{INFERCEPT} & Math & (9e-5, 6e-5) & (3.75, 1.3) \\ \cline{2-4}
        & QA & (0.69, 0.17) & (2.52, 1.73) \\ \cline{2-4}
        & VE & (0.09, 0.014) & (28.18, 15.2) \\ \cline{2-4}
        & Chatbot & (28.6, 15.6) & (4.45, 1.96) \\ \cline{2-4}
        & Image & (20.03, 7.8) & (6.91, 3.93) \\ \cline{2-4}
        & TTS & (17.24, 7.6) & (6.91, 3.93) \\ \hline
        ToolBench & - & (1.72, 3.33) & (2.45, 1.81) \\ \hline
    \end{tabular}
    \caption{API durations and number for two different datasets: INFERCEPT~\cite{abhyankarinfercept} and ToolBench~\cite{qin2023toolllm}. First part of this table is taken from INFERCEPT~\cite{abhyankarinfercept} (Table 1).}
    \label{tab:api_properties_tbl}
\end{table}

\subsubsection{Predicted API handling strategy}

Our approach integrates the selection of API handling strategies directly into the scheduling policy, determining the appropriate handling strategy before the request is processed. To do this, we first estimate the context size for batched requests, considering the memory usage of other requests that might be affected during the API handling phase. This estimation involves profiling the number of requests in a batch. API durations are predictable based on API types, as each corresponds to specific operations with known computational complexities and resource demands. For example, math APIs, which involve simple calculations, have short execution times, while image generation APIs, requiring intensive computation, have longer durations. Analysis of historical data (Table~\ref{tab:api_properties_tbl}) shows that execution times within the same API type have low variance, enabling reliable predictions. Lastly, we use a lightweight predictor (see Section~\ref{sec:evaluation} for details) to estimate the pre-API length from input prompts.

In contrast, INFERCEPT incorporates the pre-API output length and API duration for each request, also considering the impact a request has on all other requests in memory. When a request reaches the API, INFERCEPT dynamically selects a strategy to minimize memory waste based on these factors. However, its scheduling policy follows a simple FCFS, whereas our method integrates these considerations directly into a more adaptive scheduling policy.

\subsubsection{Integrating API handling strategies with scheduling}

Minimizing request completion time requires a unified scheduling method that considers both the total length of requests and their specific handling strategies during API calls. 
By knowing whether a request will Preserve, Discard, or Swap the model during an API call, the scheduler can predict the impact on system resources and the request completion time and rank the requests accordingly. For example, it may order two requests with the same total length differently because they have different handling strategies during the API call. Or it may prioritize a request with a longer total length but a more memory-friendly handling strategy during an API call over a shorter request with a handling strategy that may more negatively impact system performance, as our example showed (e.g., Figures~\ref{fig:example_opt} and~\ref{fig:example_sjf}).


\section{Design}

\begin{figure}[t]
    \centering
    \begin{subfigure}[b]{0.325\linewidth}
        \centering
        \includegraphics[width=\textwidth]{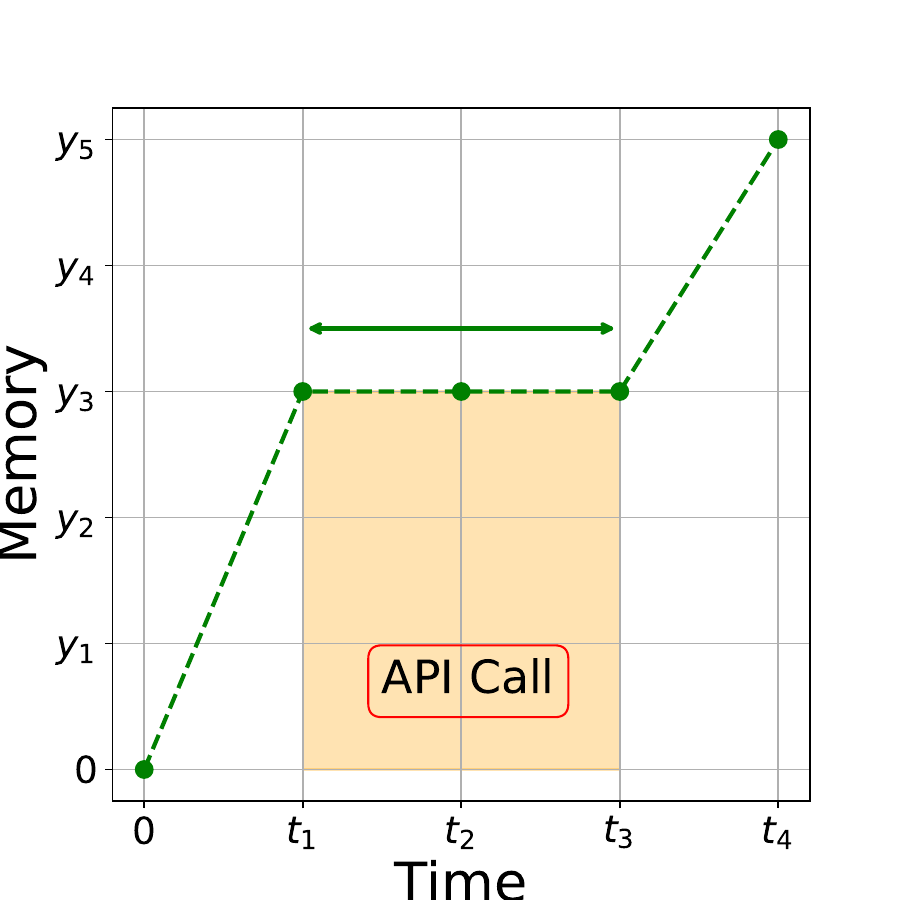}  
        \caption{Preserve}
        \label{fig:preserve_over_time}
    \end{subfigure}
    \hfill
    \begin{subfigure}[b]{0.325\linewidth}
        \centering
        \includegraphics[width=\textwidth]{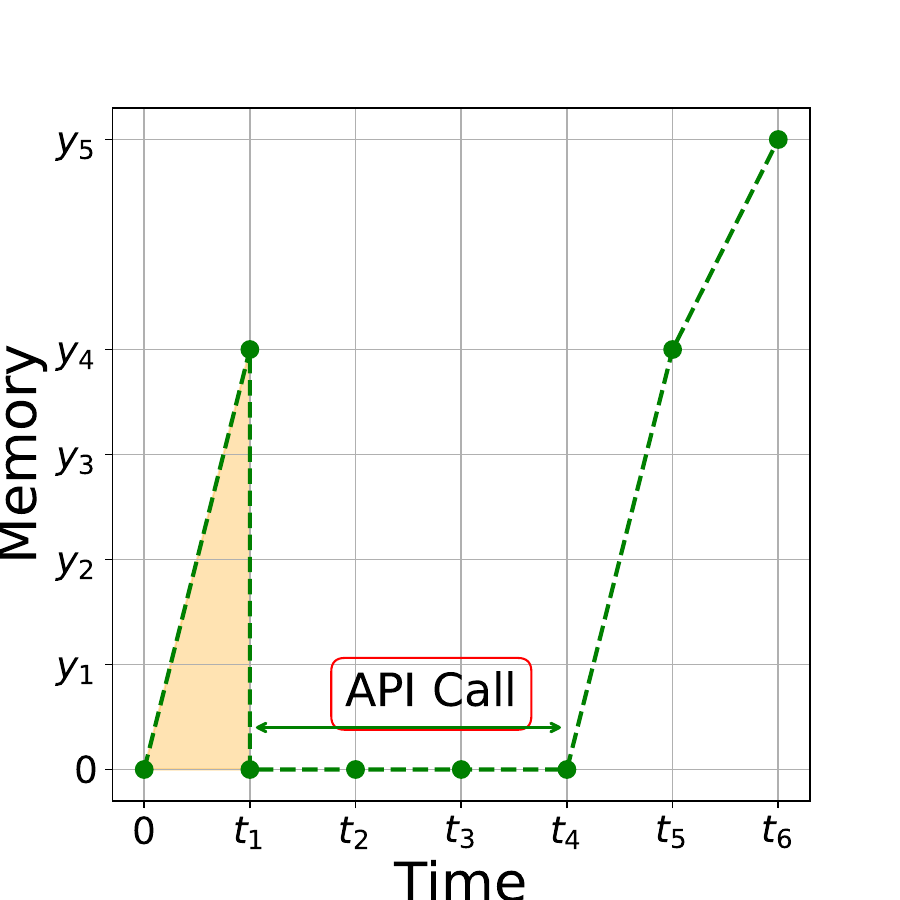}  
        \caption{Discard}
        \label{fig:recompute_over_time}
    \end{subfigure}
    \hfill
    \begin{subfigure}[b]{0.325\linewidth}
        \centering
        \includegraphics[width=\textwidth]{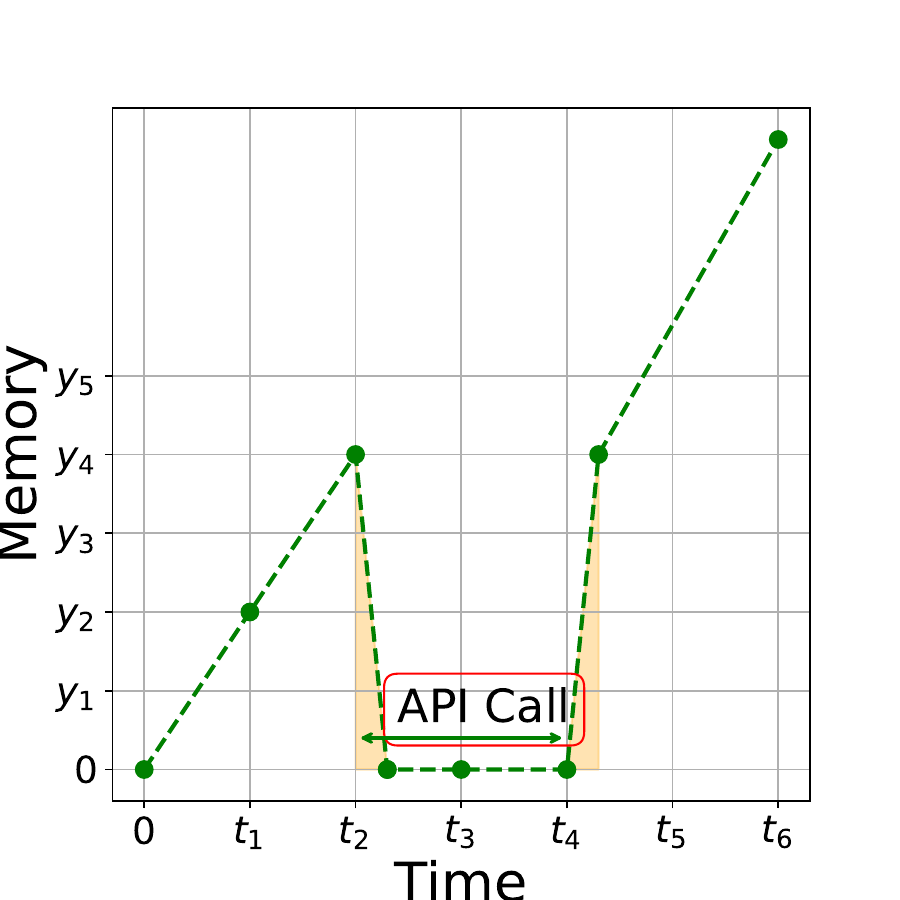}  
        \caption{Swap}
        \label{fig:swap_over_time}
    \end{subfigure}
    \caption{Memory consumption over time for a request with one API call using three memory management strategies: (1) Preserve, (2) Discard and Recompute, and (3) Swap. The highlighted area represents memory waste for one request.}
    \label{fig:memory_over_time}
\end{figure}

\subsection{System Overview}

\alg{} combines API handling strategies with scheduling policies to minimize the completion time for API-augmented requests. Using design principles in Section~\ref{sec:opportunities}, \alg{} employs three main steps to reduce LLM inference response time: predicting the pre-API output length and the duration of the API, determining the handling of requests during API calls, which aims to minimize memory waste, and, finally, proposing a scheduling policy that considers both the request length and the API handling method.

\begin{figure*}[t]
    \centering
    \includegraphics[width=0.8\linewidth]{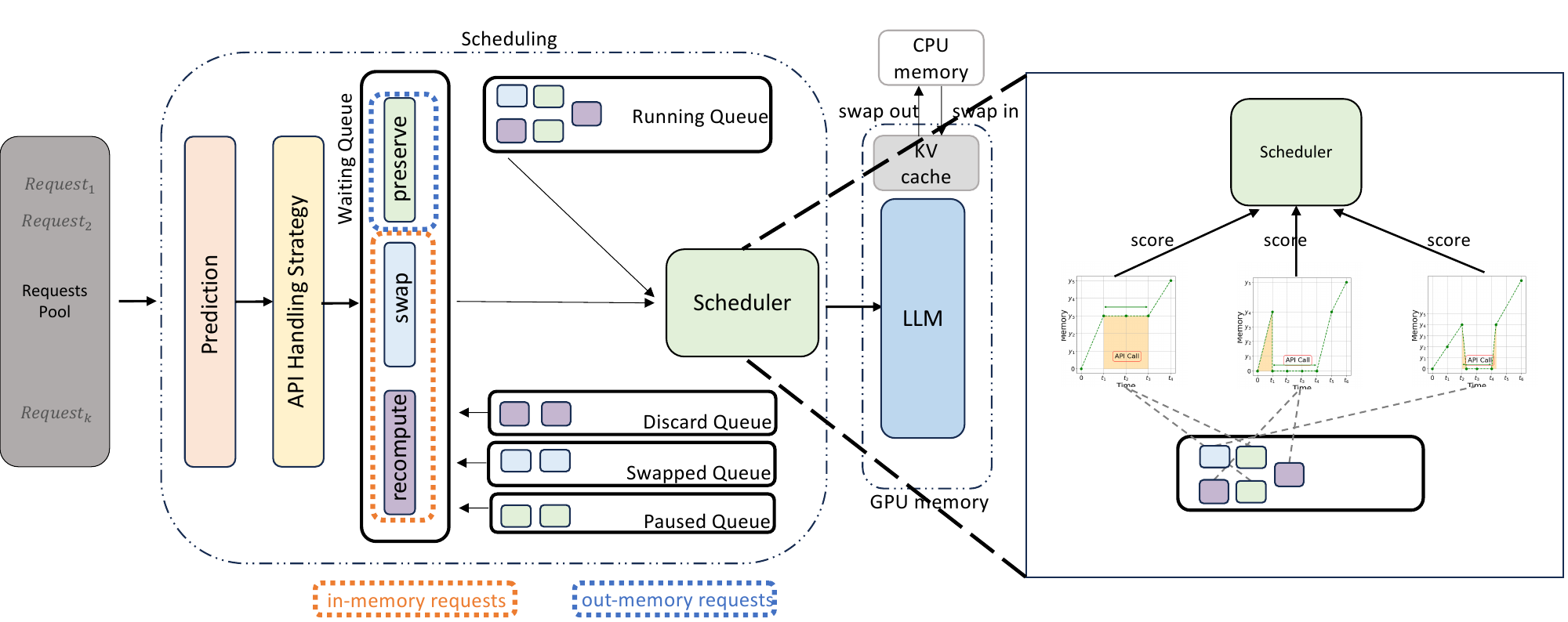}
    \caption{\alg{} architecture. \alg{} minimizes completion time for API-augmented requests through three steps: predicting pre-API output length and API properties, determining request handling strategies (preserve, discard, or swap) to minimize memory waste, and implementing a scheduling policy based on the handling method and request output length.}
    \label{fig:system_overview}
\end{figure*}

Figure~\ref{fig:system_overview} illustrates the \alg{} architecture. Users submit requests to the request pool. \alg{} predicts pre-API output length and estimates API properties (duration and response length, Table~\ref{tab:api_properties_tbl}) based on input prompts.
Using these predictions, \alg{} estimates memory consumption over time, considering this in API handling decisions and scheduling policy ranking. \alg{} determines how to handle requests during API calls to minimize memory waste, aligning with INFERCEPT (equations~\eqref{eq:waste_preserve},~\eqref{eq:waste_discard} and~\eqref{eq:waste_swap}). Each request is labeled with a handling strategy (preserve, discard, or swap). Based on this handling method and request output length, \alg{} implements a scheduling policy tailored for API-augmented requests. This policy prioritizes requests based on their memory consumption over time.
The pseudocode of the \alg{} scheduler is provided in Algorithm~\ref{alg:api_request_scheduling}.

\begin{algorithm}
\caption{\alg{} Scheduler}
\label{alg:api_request_scheduling}
\begin{algorithmic}[1]
    \State Input: Request pool $P$, predictor model $Predictor$, waiting queue $WaitingQueue$, running batch $runningBatch$, starvation threshold $StarvationT$.
    
    \While{True}
        \ForAll{$r \in P$}
            \State $predictions_r \gets Predictor(r.prompt)$
            \State $r.handling \gets HandlingStrategy(predictions_r)$
            \State $WaitingQueue.\text{put}(r)$ 
        \EndFor

        \ForAll{$r \in PQueue \cup DQueue \cup SQueue$}
            \If{$r$.\text{APIcallFinished()}}
                \State $WaitingQueue.\text{put}(r)$
            \EndIf
        \EndFor
        
        \ForAll{$r \in WaitingQueue$}
            \State $r.score \gets \text{HandlingRanking}(r)$ 
        \EndFor
        
        \State $WaitingQueue \gets \text{Sort}(WaitingQueue)$ by $r.score$

        \State $runningBatch \gets \emptyset$
        \ForAll{$r \in WaitingQueue$}
            \If{$runningBatch$ is not full}
                \State $runningBatch \gets runningBatch + r$
                \State $r.StarvationCnt \gets 0$ 
            \Else
                \State $r.StarvationCnt \gets r.StarvationCnt + 1$
            \EndIf
        \EndFor

        \ForAll{$r \in WaitingQueue$}
            \If{$r.StarvationCnt \geq StarvationT$}
                \State Place $r$ in $WaitingQueue$ head
                \State $r.StarvationCnt \gets 0$
            \EndIf
        \EndFor

        \State Remove finished requests from $WaitingQueue$
        \State Execute $runningBatch$
    \ForAll{$r \in runningBatch$}

        \If{$r.\text{encounterAPIcall()}$}
            \If{$r.handling == \textit{Preserve}$}
                \State $PQueue.\text{put}(r)$
            \ElsIf{$r.handling == \textit{Discard}$}
                \State $DQueue.\text{put}(r)$
            \ElsIf{$r.handling == \textit{Swap}$}
                \State $SQueue.\text{put}(r)$
            \EndIf
        \EndIf
    \EndFor

    \EndWhile
\end{algorithmic}
\end{algorithm}

\subsection{Predicting API Handling Strategy}


Our goal is to predict the best API handling strategy that minimizes memory waste before a request is processed, enabling the scheduler to rank each request. In the following discussion, we explain how to predict the best handling strategy for requests during API calls, assuming a single API for simplicity.
We first predict the output length and API duration. Our approach generalizes beyond the dataset by using a predictor to estimate pre-API output length based on the prompt.
Output length prediction is studied in the context of LLMs. Several works have modeled output length prediction as a classification problem, using models such as DistilBERT and OPT \cite{jin2023s, stojkovic2024dynamollm, cheng2024enabling}. These approaches categorize outputs into discrete bins to estimate lengths. Other methods use regression-based techniques to predict length \cite{qiu2024efficient, qiu2024power}. Moreover, recent work \citep{trail} demonstrates using LLM layer embeddings to predict output length. Our approach builds on these insights.

For the API duration, we leverage the fact that APIs belong to fixed classes (e.g., Math, Image), each with consistent functionality and a similar duration (latency). By extracting the API type from the prompt, we estimate the API response length using the average length from the training set for that API class. This method relies on the standardized outputs and consistent behaviors of APIs within each class.

We classify and determine how to handle requests during API calls by predicting memory waste based on pre-API and API duration estimates. INFERCEPT uses Equations~\eqref{eq:waste_preserve},~\eqref{eq:waste_discard}, and~\eqref{eq:waste_swap} to compute the memory waste. Here, we present an equivalent method to quantify memory usage (specifically calculating waste based on predictions) by considering the memory-over-time function. For simplicity, we focus on a single API, but this approach extends to multiple APIs.

Figure~\ref{fig:memory_over_time} represents the memory-over-time function and the highlighted areas in Figures~\ref{fig:preserve_over_time},~\ref{fig:recompute_over_time} and~\ref{fig:swap_over_time} represent the memory waste of a single request due to an API call. We combine this waste with our estimation of the context size for batched requests according to the setup, aligning with
Equations~\eqref{eq:waste_preserve},~\eqref{eq:waste_discard}, and~\eqref{eq:waste_swap}, respectively. After estimating the memory waste for each option, we select the strategy that minimizes the waste.
Memory consumption increases linearly with the output length until the request reaches the API call, as seen in all three cases. In Figure~\ref{fig:preserve_over_time}, the Preserve strategy keeps memory allocated throughout the API call. Thus, memory use remains constant during the call, with the shaded area indicating this idle memory. In Figures~\ref{fig:recompute_over_time} and~\ref{fig:swap_over_time}, the Discard and Swap strategies release memory during the API call, resulting in zero memory usage while waiting for the API response. In Figure~\ref{fig:recompute_over_time}, the Discard approach recomputes the first part after the API, while the Swap approach (Figure~\ref{fig:swap_over_time}) shows a delay before the memory is fully released (swapped out) and then restored (swapped in) after the API call completes. Swapping in causes a spike in memory consumption.

Our insight is that evaluating memory usage by integrating the memory-over-time function offers a more accurate measure of resource consumption than relying on instantaneous memory values. The integral of memory versus time provides the total memory consumption for a request, accounting for the API's impact on memory usage. This approach incorporates output length, API duration, and how the request is handled during the API call. Instantaneous memory measurements fail to reflect how long memory resources are occupied, which is particularly important in the decoding phase of LLMs where the system is memory-bound. It is not just the amount of memory a request consumes at a particular moment but also how long that memory remains in use. A strategy that uses more memory for a shorter period can be more efficient than one that uses less memory but occupies it longer. Integrating memory over time captures memory waste across different strategies during API calls.


\textbf{Multi-API.}
To generalize to requests involving multiple APIs, we break down each request into segments, each ending with a single API call. After an API call completes, the request re-enters the system for further processing and is treated as a new request focused on the subsequent API call. At this point, we classify the request based on the characteristics of the current API.
For example, suppose a job involves initial processing, followed by two API calls interleaved with additional processing phases. We divide this job into segments, each consisting of a processing phase and a single API call at the end. We estimate the returned token length in each segment based on the specific API call.
While this approach does not account for the cumulative memory consumption of the entire job, predicting the total number of API calls and their combined resource usage beforehand is challenging. This segmentation aligns with INTERCEPT, which handles multi-API requests by processing jobs incrementally as they reach each API.

\textbf{Effect of Mispredictions.}
Mispredictions are to be expected. Small mispredictions in API duration or output size will typically have a small effect; indeed, they may not change the overall ranking of jobs.
Mispredicting a short API or output as long may have a large effect on that particular job, but does not typically harm other jobs in the system. (See related results in \cite{Mitzenmacher21}.)
A long-running API call incorrectly predicted as short may lead the system to select a memory-wasteful strategy, such as preserving the request in memory during the API call. This unnecessary memory consumption may limit the system's ability to process additional requests and reduce overall throughput. A request with a long output misclassified as short may cause head-of-line blocking and delay other requests. The main overhead in this scenario is increased latency.


\subsection{Scheduling Policy}
Traditional job scheduling typically assumes that job completion times are either completely unknown—making First-Come, First-Served (FCFS) a natural strategy—or known in advance or predictable, enabling size-based policies like Shortest Job First (SJF) and Shortest Remaining Process Time (SRPT) to minimize average response time.

Integrating API calls into the output response increases memory consumption, which may degrade performance due to memory constraints.
In LLM systems, when memory is full, jobs are either discarded and recomputed when memory becomes available, or KV cache entries are swapped from the GPU to the CPU. Both approaches impact response time: discarding requires recomputation, and swapping interrupts the model's forward pass, causing delays for the entire batch.
Intuitively, requests should be ranked based on their memory consumption. Without API calls, ranking based on memory consumption aligns with ranking based on service time (or request length), as memory consumption has a linear relationship with request length. With API calls, this relationship breaks, as requests are handled differently according to the strategy that minimizes memory consumption during the API. Consider Figure~\ref{fig:memory_over_time}, which shows memory over time; we consider the area (integral) as a rank function of a request and select the function based on the predicted handling strategy during the API call. For example, the memory over time function matches Figure~\ref{fig:preserve_over_time} for the predicted preserve strategy for a request. This approach incorporates the strategy of handling requests during API within the scheduling policy and provides a way to compare and rank requests among different handling strategies. Referencing Figure~\ref{fig:example}, consider memory consumption over time. Among the three requests, $R_3$ consumes the least memory and should be prioritized, followed by $R_2$. $R_1$ consumes the most memory due to its length, API duration, and the preserve handling strategy, so it should be scheduled last.

Our scheduling strategy uses iteration-level scheduling~\cite{yu2022orca}, where the scheduler ranks requests at the end of each iteration. We use a selective score update mechanism to reduce the overhead of frequent ranking. For example, for datasets with long-running requests, frequent score updates are unnecessary; instead, we cache their scores and refresh them at predefined intervals. This balances ranking accuracy with the computational costs of maintaining updated scores.

\subsection{Starvation Prevention}

Scheduling policies can cause certain requests to experience long wait times, leading to high tail latency, a form of starvation that degrades system performance and user experience. This issue arises when longer or resource-intensive requests are continually deferred in favor of shorter ones, exacerbating tail latency. Our memory-focused scheduling policy alone does not detect and mitigate starvation, which can result in extended wait times and reduced fairness.
To solve this, we have implemented a starvation prevention mechanism to improve the scheduler's tail latency using a per-request counter. The counter increments when a request remains in the waiting queue for a new iteration. Upon reaching a predefined threshold, \alg{} tags the request as starving and prioritizes it by placing it at the head of the scheduled requests for the current iteration. The relative order of prioritized and non-prioritized requests is maintained according to \alg{}'s ranking decisions. Prioritization continues until request completion, avoiding memory waste from preempted (half-finished) requests. If the request has not been prioritized and encounters API calls or is scheduled, the counter resets to 0. Parameter experiments led us to set the predefined threshold at $100$ (testing with the datasets in Section~\ref{sec:evaluation}). This value effectively filters out requests that complete within a reasonable time while identifying starving ones. The starvation prevention mechanism activates only when \alg{} initially schedules the request to the running queue, thus leveraging \alg{}'s sorting decisions.

\section{Implementation}
We implement \alg{} on top of vLLM~\cite{kwon2023efficient}, a state-of-the-art LLM inference system.
We enable the score update mechanism only on the ToolBench dataset with an interval of ten, while disabling it for other datasets where scheduling overhead is not a bottleneck.
To implement the prediction mechanism, we use the OPT-125M language model~\cite{zhang2022opt}, a transformer-based model developed by Meta. With 125 million parameters and support for a context length of 2048 tokens, OPT-125M can effectively handle datasets with long contexts, such as the multi-API dataset. Although smaller than many larger language models, OPT-125M delivers strong language generation capabilities. 
Our approach utilizes the embeddings generated by OPT-125M during the initial processing of input prompts. After tokenizing the input and processing it through the model's layers, we extract the final token's embedding, which is then fed into a linear classifier. This classifier assigns the input to one of 50 bins, each representing a range of 10 tokens, and is trained using cross-entropy loss.

The model estimates the completion length for each prompt based on learned representations from the ToolBench dataset \cite{qin2023toolllm}, which involves complex conversations with API interactions.
We train the model using an 80-20 split for training and validation, classifying output lengths into bins. We apply this model specifically to the ToolBench dataset because the other dataset already includes detailed output length information, making prediction unnecessary in that case. \alg{} is evaluated using the test portion of the ToolBench data to ensure accuracy.


\section{Evaluation}
\label{sec:evaluation}

In this section, we first present end-to-end experiments demonstrating the overall performance improvements of \alg{} compared to INFERCEPT and vanilla vLLM on two different-sized LLM models. Next, we evaluate \alg{}'s design choices, highlighting the effectiveness of each component. Finally, we analyze the prediction component and the impact of mispredictions on \alg{}'s performance.

\subsection{Methodology}

\textbf{LLM models.} We use the 6B-parameter GPT-J model (which we denote by GPT-J 6B), and the 13B-parameter Vicuna model (Vicuna 13B). Both were also used by INFERCEPT.



\textbf{Testbed.}
For the experiments, we used a machine with dual AMD EPYC 7313 CPUs (16 cores per CPU, totaling 64 threads), 503 GB of RAM, and two NVIDIA A100 GPUs with 80 GB memory each connected via NVLink. We manually limited the maximum memory usage of each GPU to 40 GB to emulate the setup used in INFERCEPT's experiments, which we inferred to involve A100 GPUs with 40 GB memory based on their use of AWS machines.

\textbf{Datasets.}
We evaluate our system using two distinct datasets. The first, similar to the one used in INFERCEPT, includes arithmetic tasks, knowledge-based question answering, multi-step chatbot dialogues, and interactions in an embodied virtual environment. This dataset includes API execution times, frequencies, and output length.
The second data set is ToolBench~\cite{qin2023toolllm} is an instruction-tuning dataset tailored for tool-use tasks, comprising over 16,000 real-world APIs across 49 categories. It encompasses both single-API and multi-API scenarios, containing only prompts and API call types. We use this dataset to predict output length, API duration, and API response length.

\textbf{Metrics.}
To evaluate \alg{}, we measure end-to-end latency, defined as the time from when a request is submitted to the system until its completion, time-to-first-token (TTFT) (reflecting system responsiveness), and throughput (indicating request generation speed). For each metric, we report the mean and 99th percentile (P99).

\textbf{Baselines.}
As baselines, we compare \alg{} with both vanilla vLLM and INFERCEPT.

\begin{figure*}[ht!]
    \centering
    \begin{tikzpicture}
        \node at (5.5, 12) {\textbf{GPT-J 6B}};
        \node at (14.5, 12) {\textbf{Vicuna 13B}};
        
        \node[rotate=90] at (0.7, 10) {\textbf{Single}};
        \node[rotate=90] at (0.7, 8) {\textbf{Multi}};
        \node[rotate=90] at (0.7, 6) {\textbf{ToolBench}};

        \draw[dashed, gray, very thin] (9.8, 11) -- (9.8, 5);

        \node at (10, 4.5) {
            \begin{minipage}{0.4\textwidth}
                \centering
                \includegraphics[width=\textwidth]{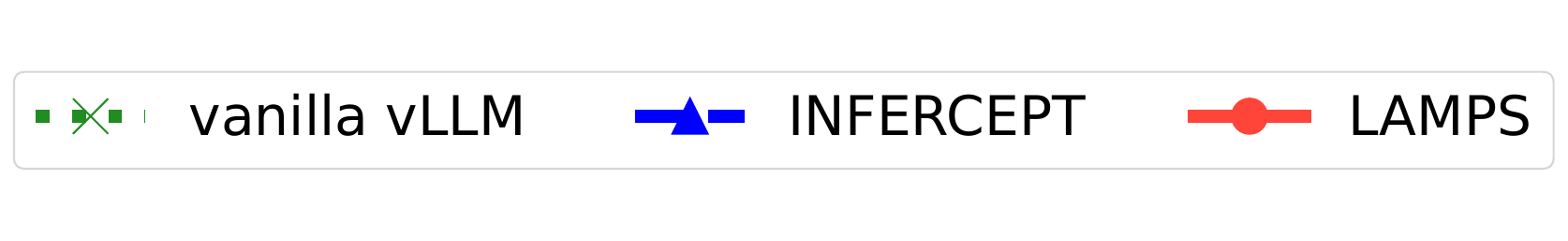}
            \end{minipage}
        };
        \node at (5.5, 10) {
            \begin{minipage}{0.5\linewidth}
                \centering
                \includegraphics[width=0.22\linewidth]{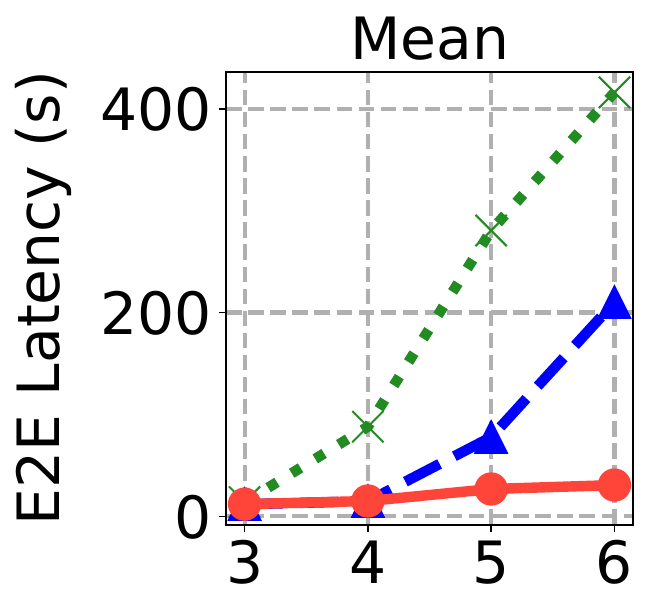}
                \includegraphics[width=0.22\linewidth]{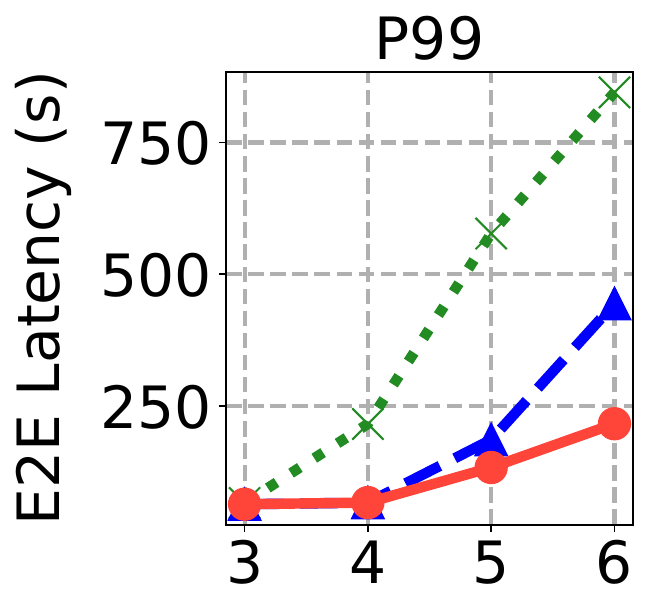}
                \includegraphics[width=0.22\linewidth]{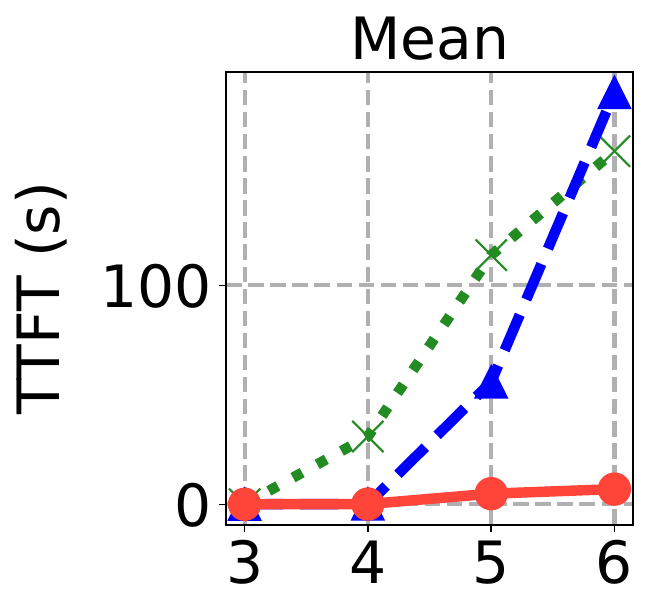}
                \includegraphics[width=0.22\linewidth]{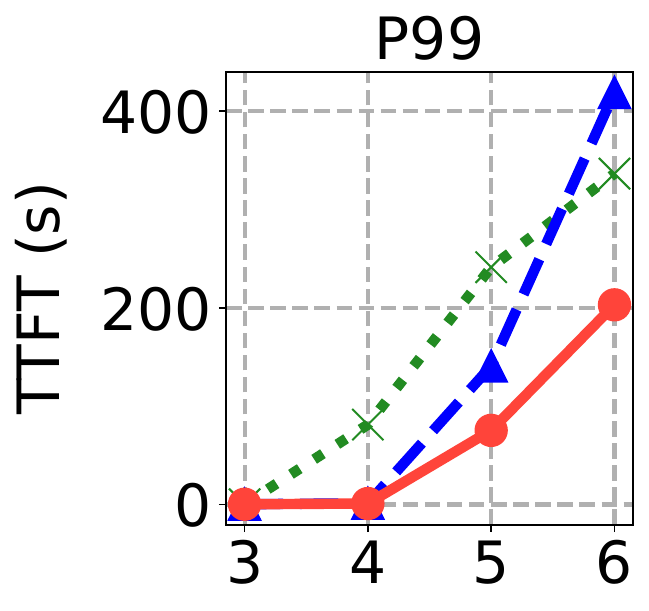}
            \end{minipage}
        };
        \node at (14, 10) {
            \begin{minipage}{0.5\linewidth}
                \centering
                \includegraphics[width=0.22\linewidth]{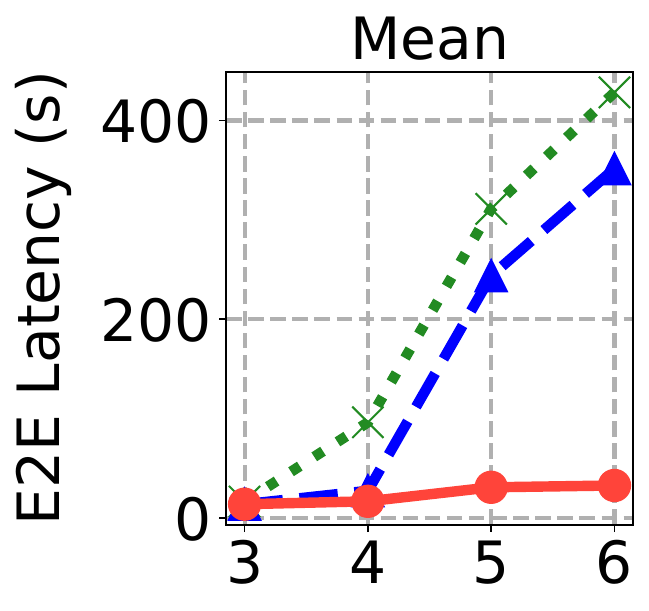}
                \includegraphics[width=0.22\linewidth]{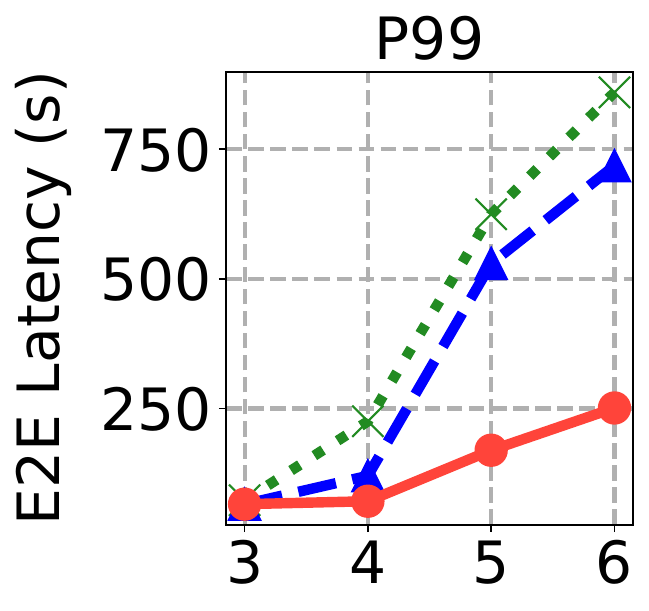}
                \includegraphics[width=0.22\linewidth]{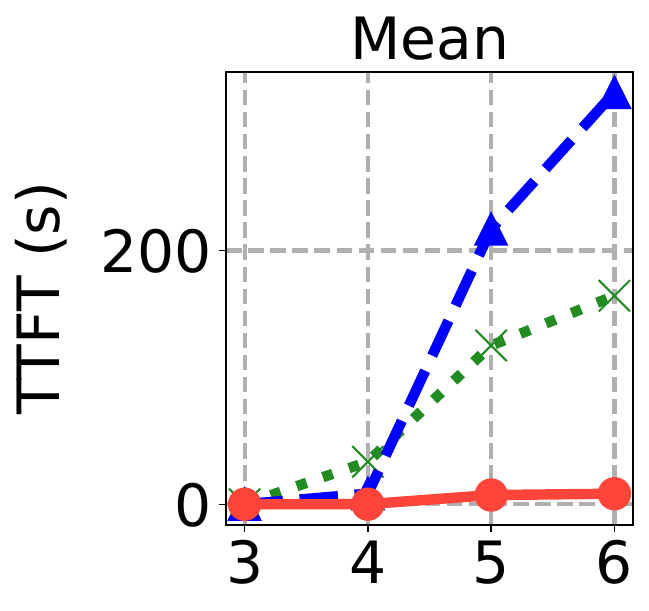}
                \includegraphics[width=0.22\linewidth]{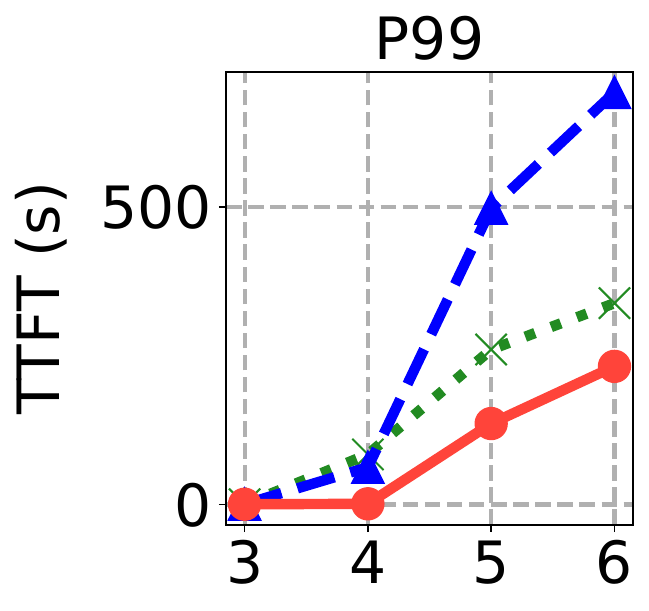}
            \end{minipage}
        };

        \node at (5.5, 8) {
            \begin{minipage}{0.5\linewidth}
                \centering
                \includegraphics[width=0.22\linewidth]{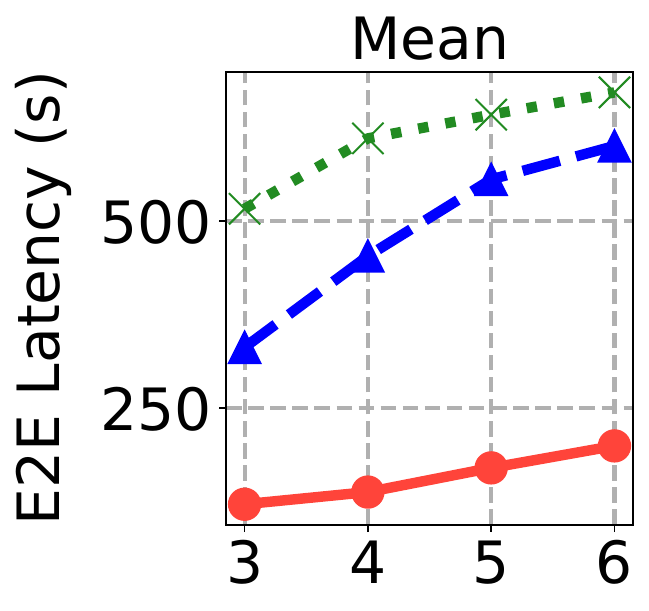}
                \includegraphics[width=0.22\linewidth]{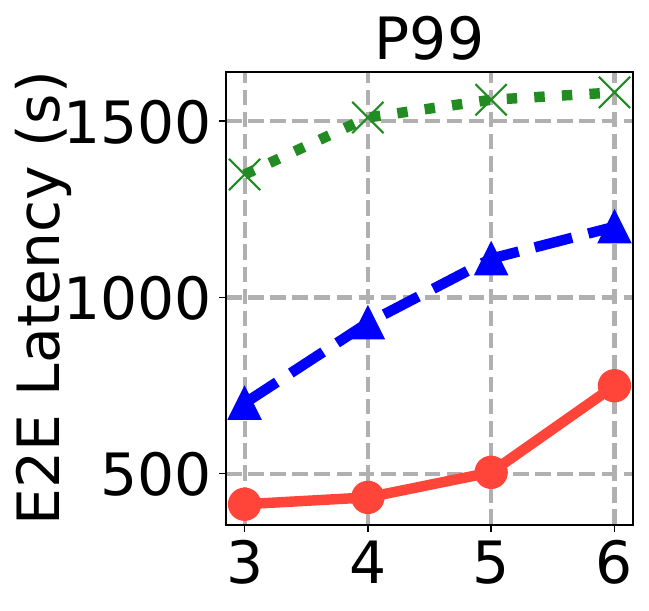}
                \includegraphics[width=0.22\linewidth]{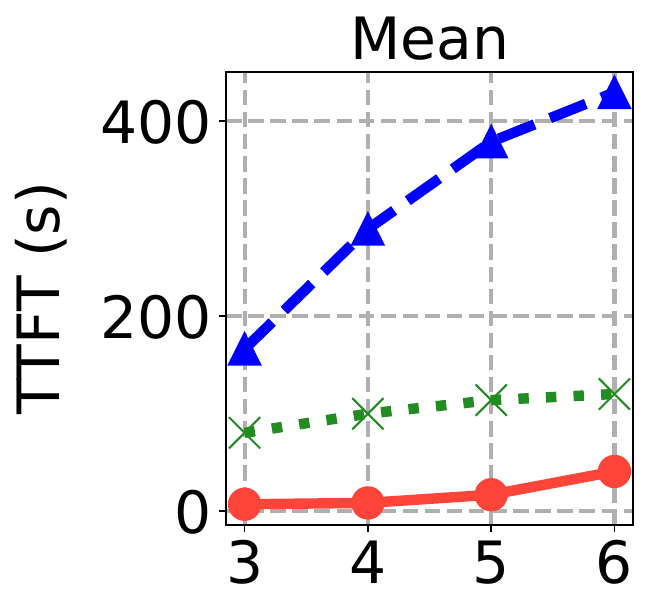}
                \includegraphics[width=0.22\linewidth]{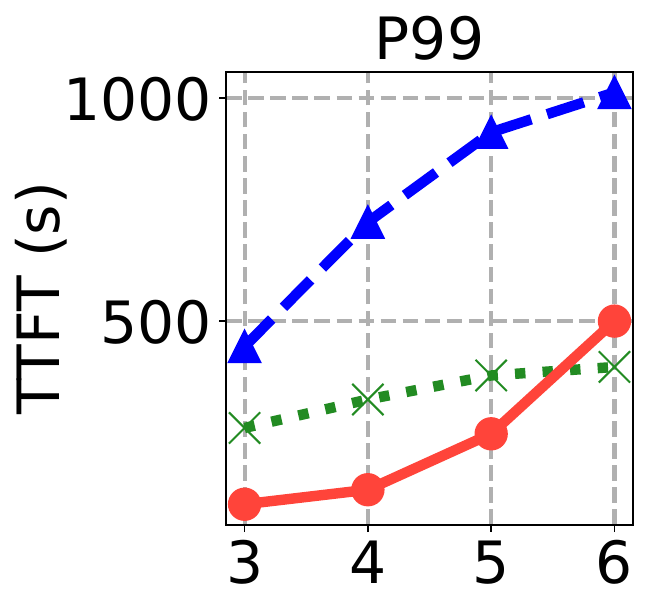}
            \end{minipage}
        };
        \node at (14, 8) {
            \begin{minipage}{0.5\linewidth}
                \centering
                \includegraphics[width=0.22\linewidth]{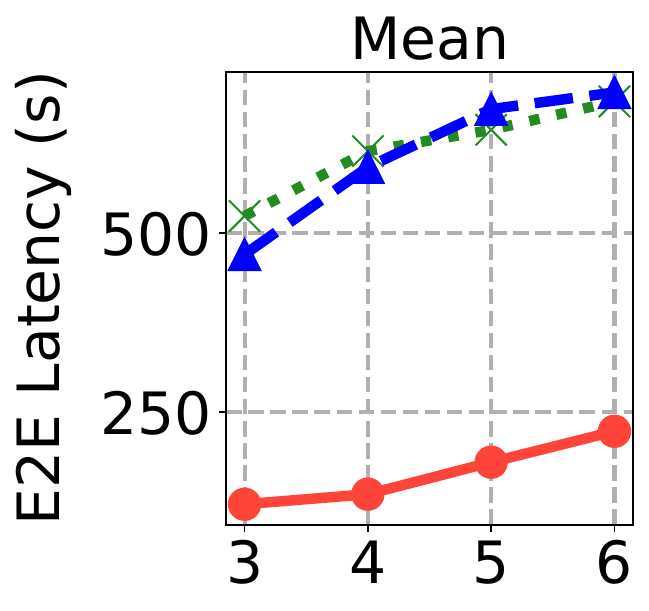}
                \includegraphics[width=0.22\linewidth]{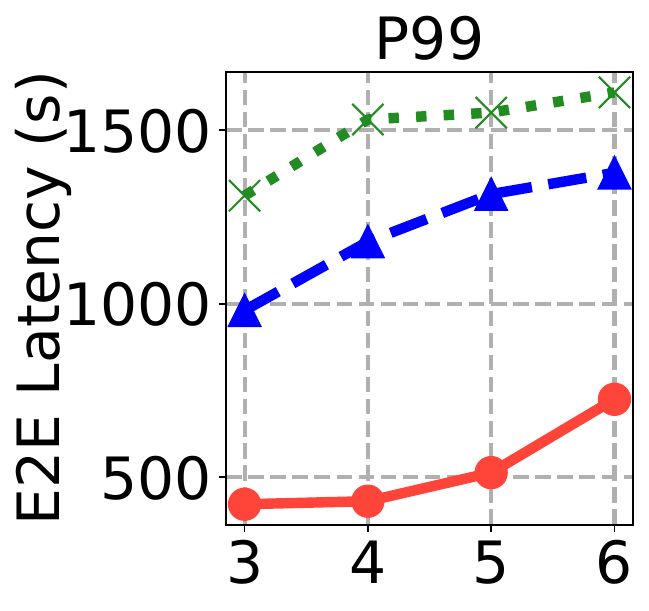}
                \includegraphics[width=0.22\linewidth]{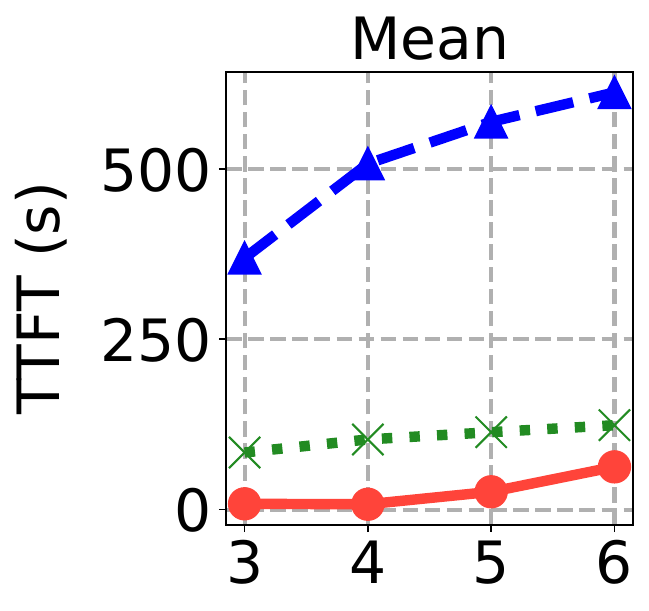}
                \includegraphics[width=0.22\linewidth]{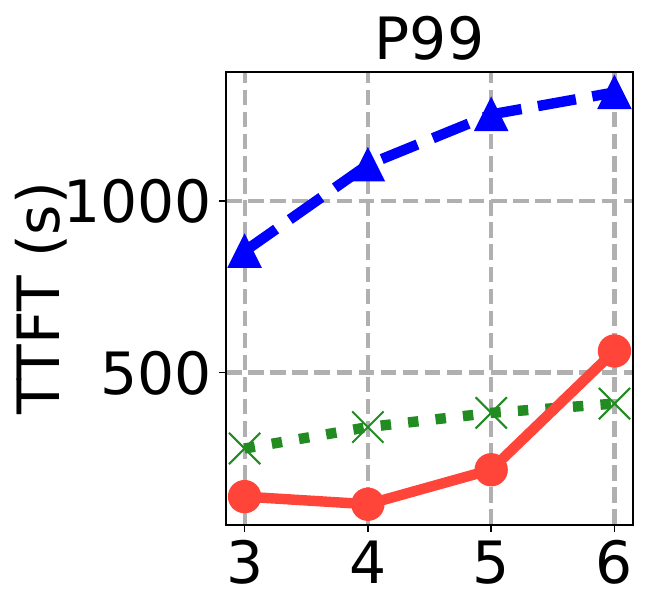}
            \end{minipage}
        };

        \node at (5.5, 6) {
            \begin{minipage}{0.5\linewidth}
                \centering
                \includegraphics[width=0.22\linewidth]{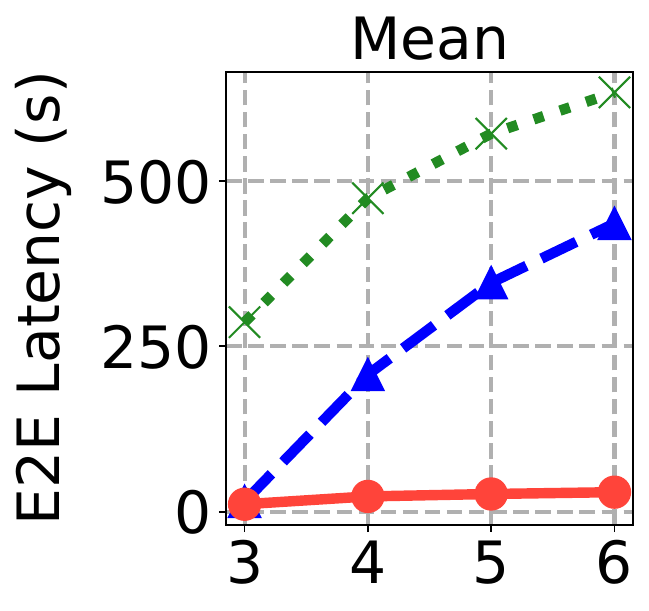}
                \includegraphics[width=0.22\linewidth]{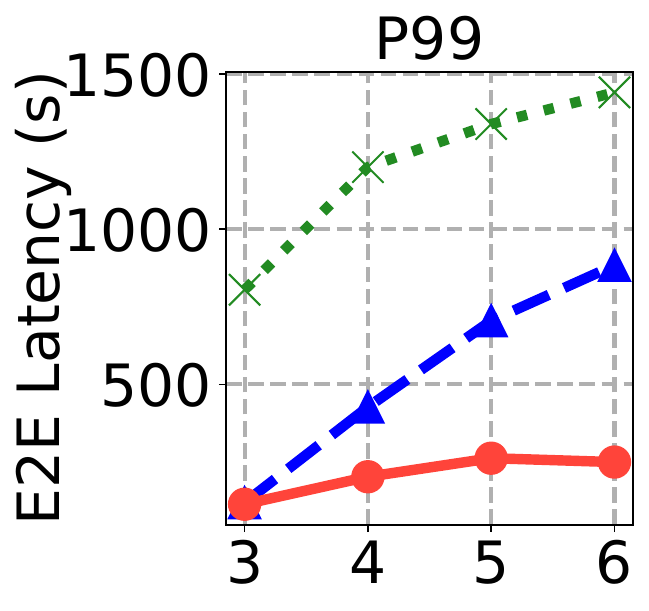}
                \includegraphics[width=0.22\linewidth]{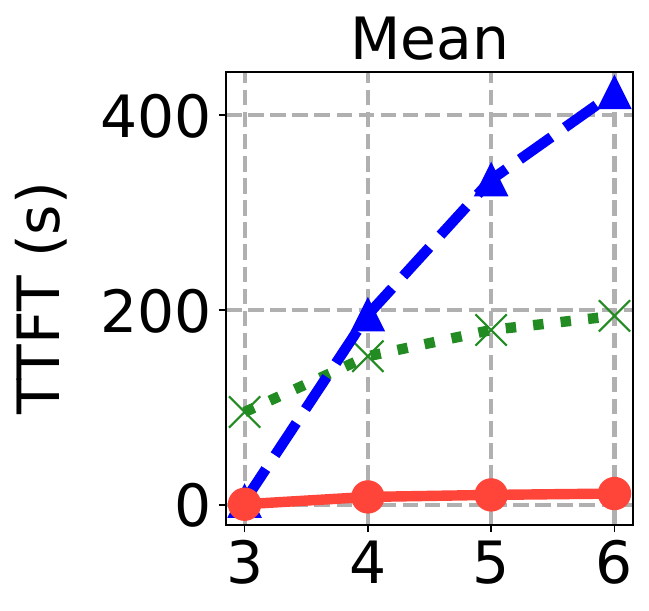}
                \includegraphics[width=0.22\linewidth]{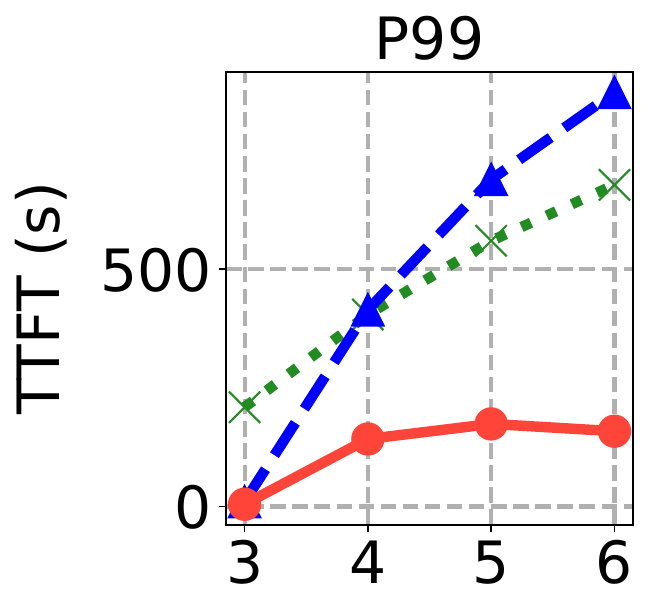}
            \end{minipage}
        };
        \node at (14, 6) {
            \begin{minipage}{0.5\linewidth}
                \centering
                \includegraphics[width=0.22\linewidth]{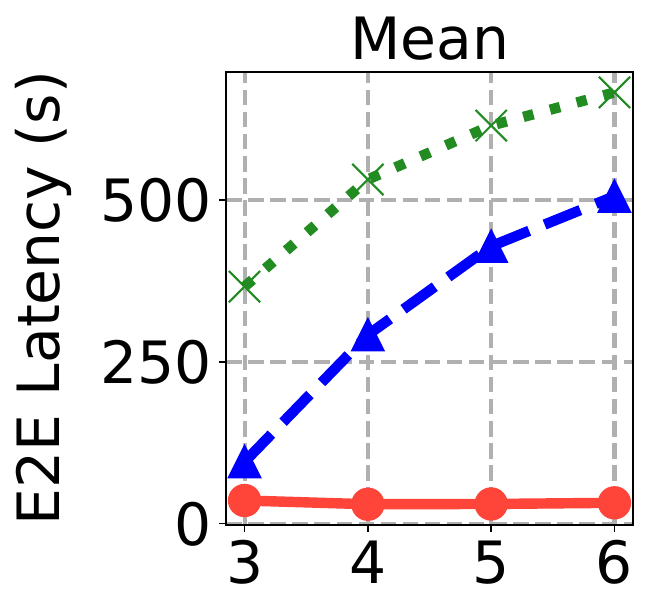}
                \includegraphics[width=0.22\linewidth]{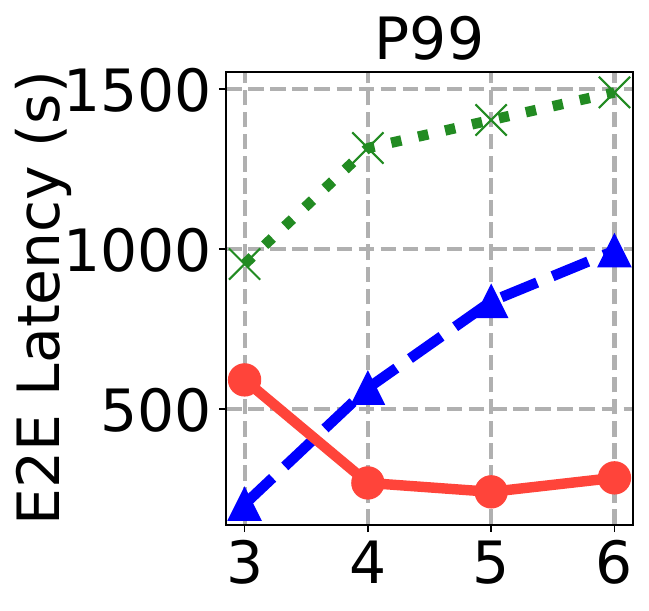}
                \includegraphics[width=0.22\linewidth]{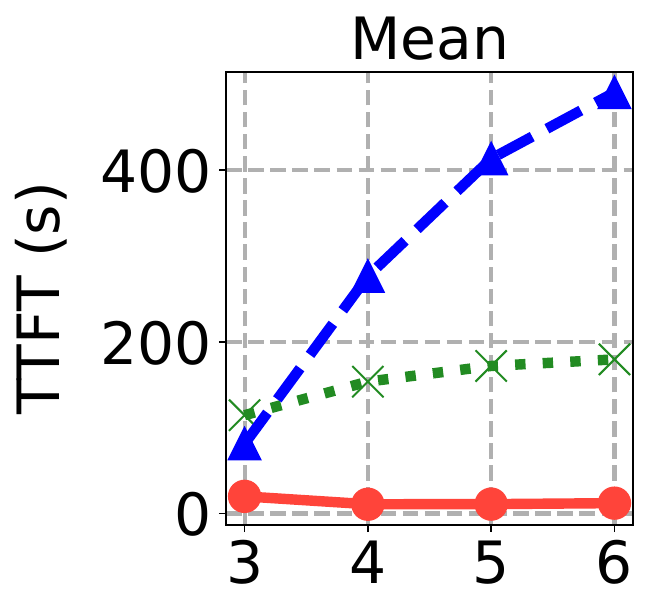}
                \includegraphics[width=0.22\linewidth]{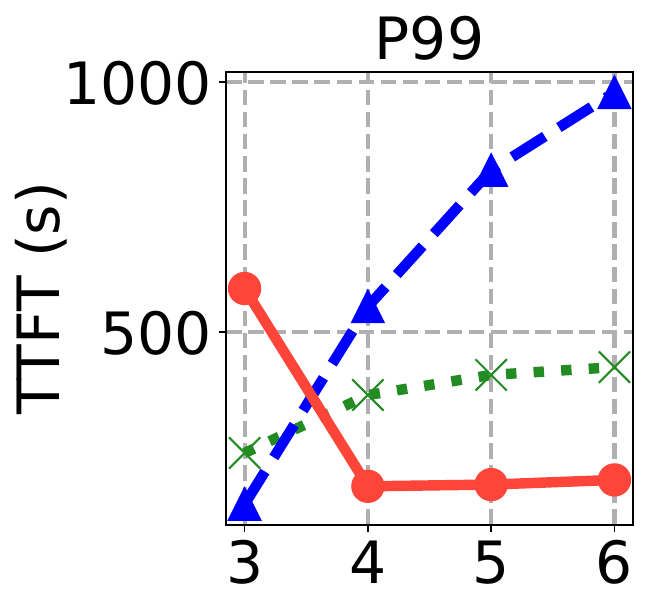}
            \end{minipage}
        };
    \end{tikzpicture}
    \caption{End-to-end performance (mean and P99 of latency and TTFT) as a function of request arrival rate when serving GPT-J 6B and Vicuna 13B using different datasets (single-API, multi-API, ToolBench).}
    \label{fig:perf_arrival_rate}
\end{figure*}

\begin{figure*}[t]
    \centering
    \begin{subfigure}{0.2\textwidth}
        \centering
        \includegraphics[width=\textwidth]{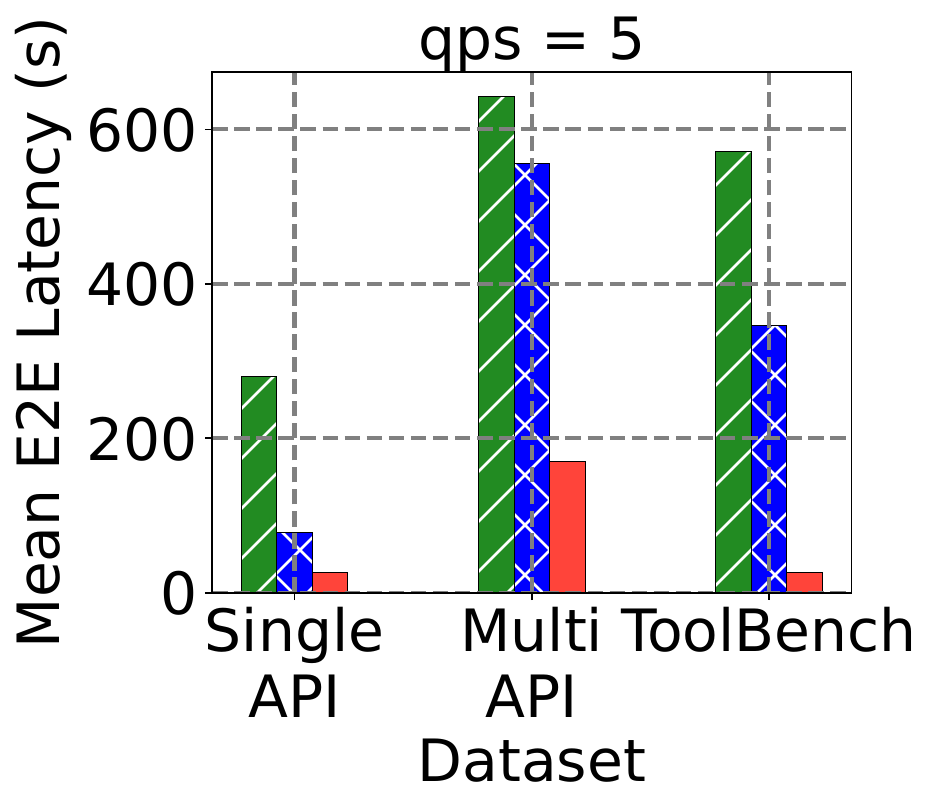}
        \caption{GPT-J 6B}
        \label{fig:subfig1}
    \end{subfigure}
    \begin{subfigure}{0.2\textwidth}
        \centering
        \includegraphics[width=\textwidth]{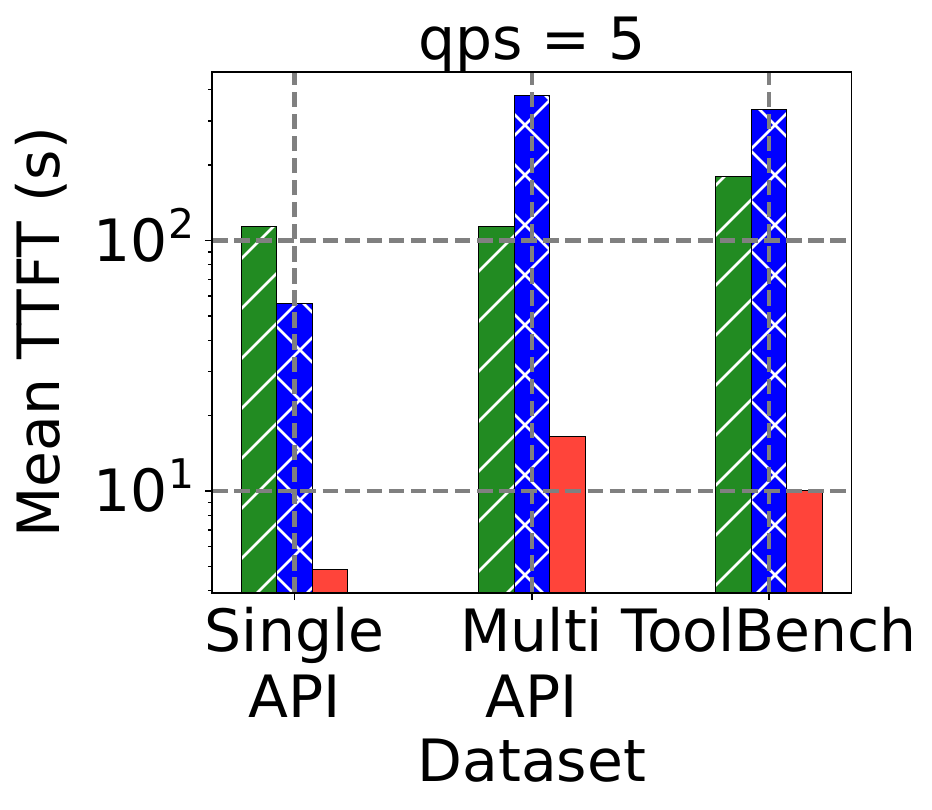}
        \caption{GPT-J 6B}
        \label{fig:subfig2}
    \end{subfigure}
    \begin{subfigure}{0.2\textwidth}
        \centering
        \includegraphics[width=\textwidth]{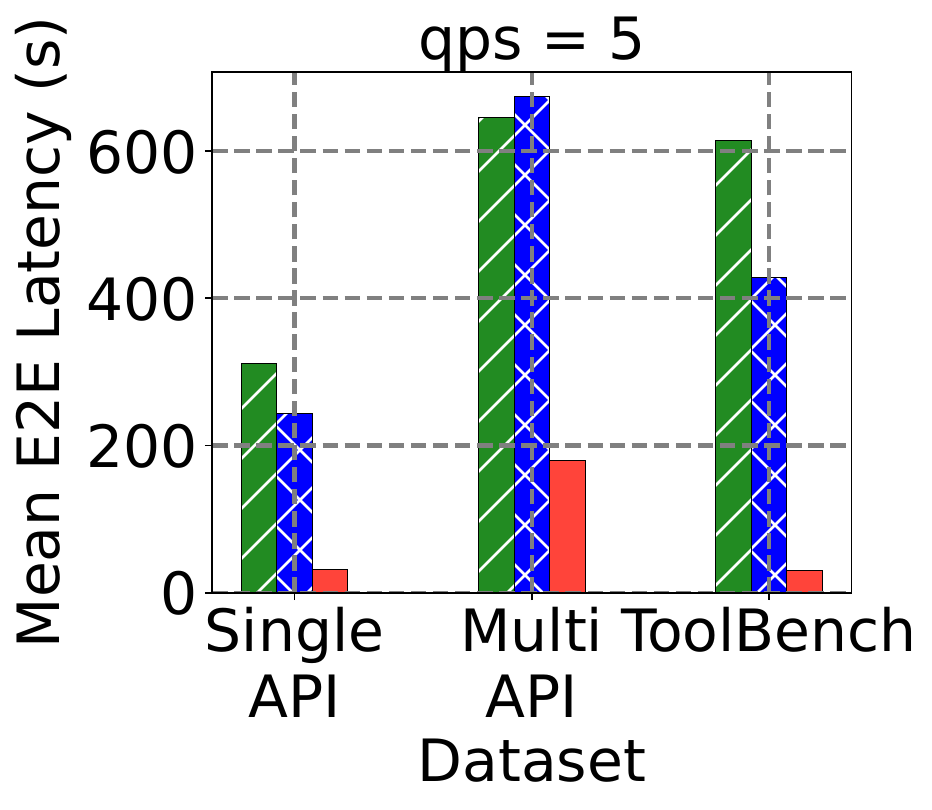}
        \caption{Vicuna 13B}
        \label{fig:subfig4}
    \end{subfigure}
    \begin{subfigure}{0.2\textwidth}
        \centering
        \includegraphics[width=\textwidth]{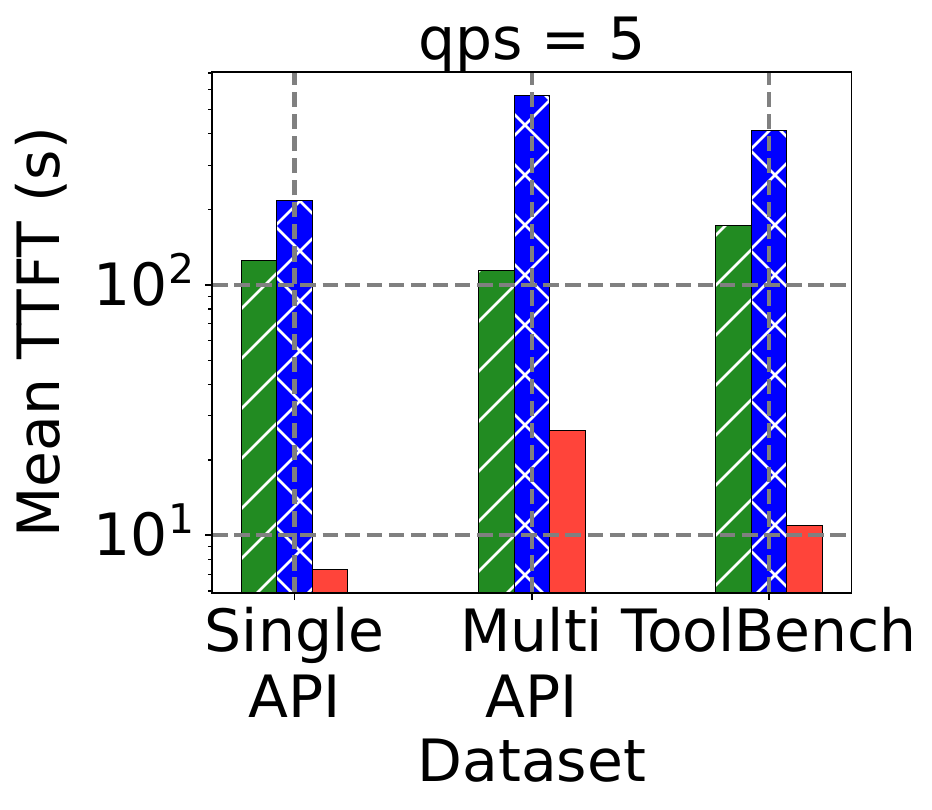}
        \caption{Vicuna 13B}
        \label{fig:subfig5}
    \end{subfigure}
    
    \includegraphics[width=0.4\textwidth]{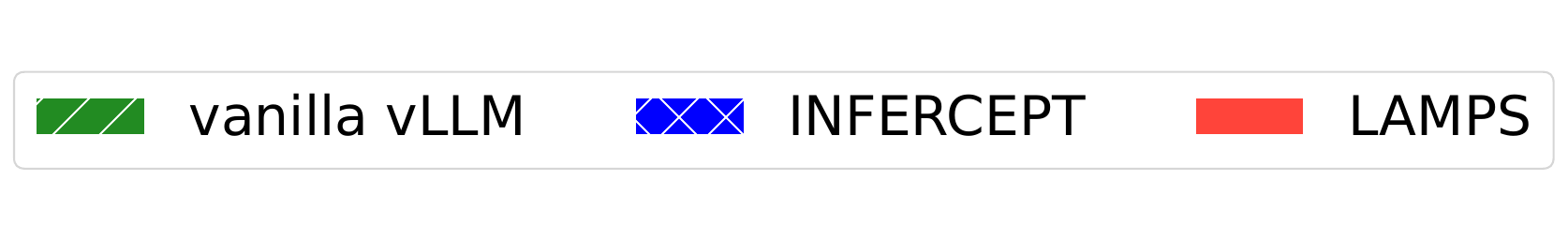}
    \caption{End-to-end performance (mean and P99 of latency and TTFT) of single-API, multi-API, ToolBench datasets when the request arrival rate is fixed to five when serving GPT-J 6B and Vicuna 13B.}
    \label{fig:perf_vs_dataset}
\end{figure*}

\subsection{End-to-end Performance}

\textbf{End-to-end latency and TTFT vs. request rate.}
Figure~\ref{fig:perf_arrival_rate} shows how varying the request arrival rate affects the mean and P99 of end-to-end latency and TTFT across three datasets: (1) a single-API dataset (a subset of the INFERCEPT dataset containing only a single API), (2) the full INFERCEPT dataset, and (3) the ToolBench dataset. We evaluated these metrics using the LLMs GPT-J 6B and Vicuna 13B.

\textbf{GPT-J 6B results.}
\alg{} shows clear performance gains over vLLM and INFERCEPT in mean TTFT and end-to-end latency across all tested datasets. On the single-API dataset at a request rate of 3, \alg{} reduces mean TTFT by 4.61\% compared to INFERCEPT and 22.86\% compared to vLLM. Mean end-to-end latency increases slightly by 0.78\% against INFERCEPT but drops by 14.48\% compared to vLLM. At higher rates, such as 5, \alg{} further reduces mean TTFT by 91.27\% over INFERCEPT and 95.71\% over vLLM, with latency reductions of up to 65.51\% and 90.44\%, respectively. For the multi-API dataset, \alg{} achieves 95.93\% TTFT reduction and 63.32\% latency improvement over INFERCEPT at a rate of 3. On ToolBench, \alg{} reduces mean TTFT by 87.04\% compared to INFERCEPT and 99.51\% compared to vLLM, with a 27.24\% latency reduction compared to INFERCEPT and 96.07\% compared to vLLM at a request rate of 3.


\textbf{Vicuna 13B results.}
\alg{} consistently outperforms vLLM and INFERCEPT in TTFT and end-to-end latency across all datasets. On the single-API INFERCEPT dataset, at a request rate of 3, \alg{} reduces mean TTFT by approximately 4.78\% compared to INFERCEPT and 18.65\% compared to vLLM, with mean end-to-end latency reductions of around 0.24\% and 15.73\%, respectively. For higher request rates (e.g., 4), \alg{} achieves significantly greater improvements, reducing mean TTFT by over 98\% compared to baselines and end-to-end latency by up to 82\%. On the multi-API dataset, \alg{} achieves similar gains, with TTFT reductions of over 89\% and mean end-to-end latency improvements of up to 78\% at 4. In the ToolBench dataset, at a request rate of 3, \alg{} reduces mean TTFT by 75.46\% and end-to-end latency by 62.61\% compared to INFERCEPT while also achieving improvements over vLLM by up to 90.25\% in mean latency.


At low request rates in a single-API dataset, the performance gap between \alg{} and the baseline methods is small, as FCFS and size-based policies perform similarly under less system pressure. However, as the arrival rate increases or with a multi-API dataset, the head-of-line blocking problem worsens with FCFS. For P99 latency and TTFT, while \alg{} shows a higher growth rate in these metrics at higher request rates, its absolute TTFT values remain lower than the baselines. This is because \alg{} prioritizes low-memory requests, leading to increased tail latency and TTFT at higher rates due to more pronounced starvation effects.

Figure~\ref{fig:perf_vs_dataset} shows the mean end-to-end latency and mean time-to-first-token (TTFT) for different datasets with a fixed arrival rate of five. All systems exhibit a similar pattern across the datasets, with higher latency and TTFT for the Multi API and ToolBench datasets than the Single API dataset. However, \alg{} consistently achieves lower latency and TTFT across all datasets.

\begin{figure}[t]
    \centering
    \begin{subfigure}[b]{0.325\linewidth}
        \centering
        \includegraphics[width=\textwidth]{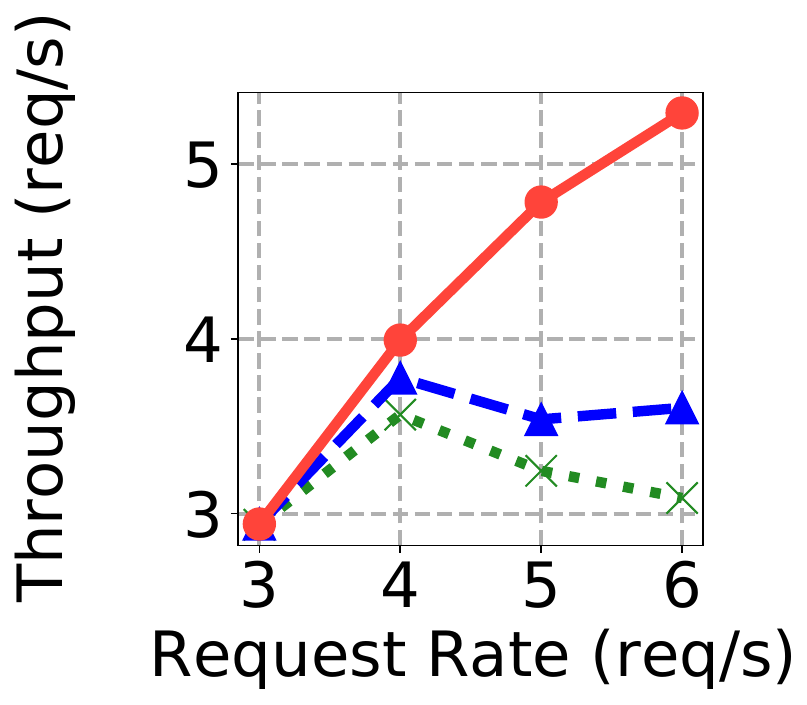}  
        \caption{Single-API}
        \label{fig:throughput1}
    \end{subfigure}
    \hfill
    \begin{subfigure}[b]{0.325\linewidth}
        \centering
        \includegraphics[width=\textwidth]{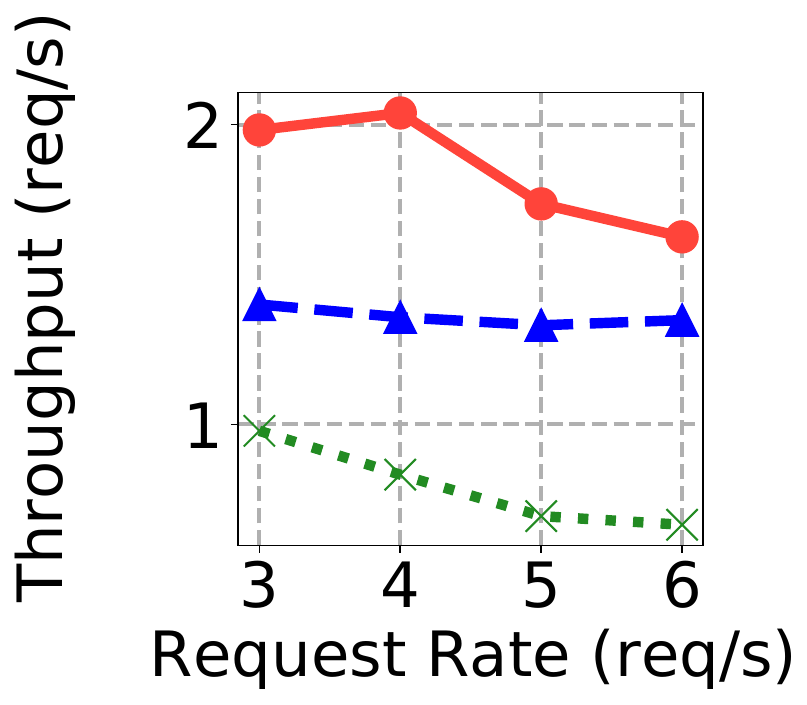}  
        \caption{Multi-API}
        \label{fig:throughput2}
    \end{subfigure}
    \hfill
    \begin{subfigure}[b]{0.325\linewidth}
        \centering
        \includegraphics[width=\textwidth]{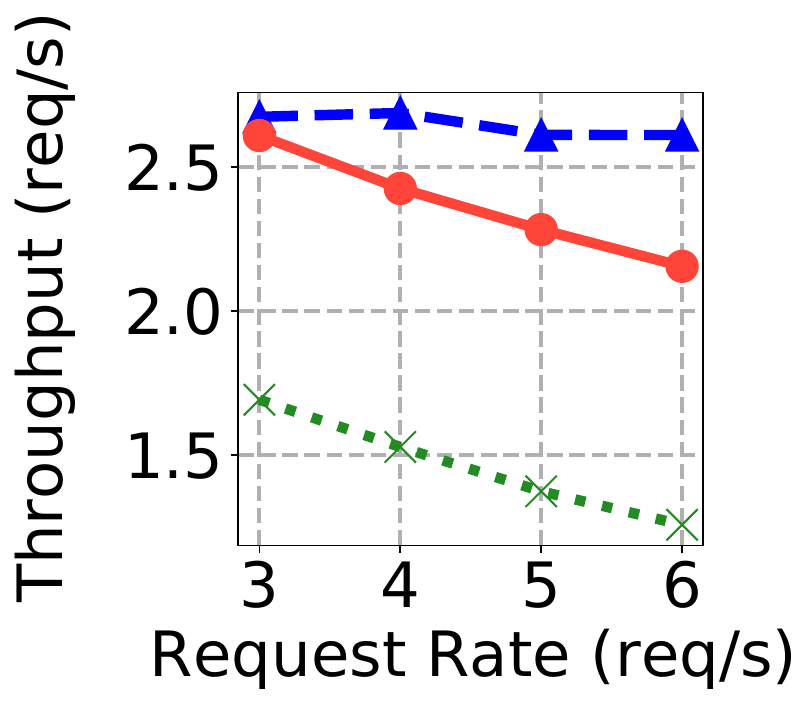}  
        \caption{ToolBench}
        \label{fig:throughput3}
    \end{subfigure}
    \includegraphics[width=0.7\linewidth]{eval_graphs/arxiv_legends/for_performance/strategy_legend.pdf}
    \caption{Throughput as a function of request arrival rate with Vicuna 13B using the different datasets.}
    \label{fig:throughput}
\end{figure}

\textbf{Throughput vs. request rate.}
Figure~\ref{fig:throughput} shows throughput vs. request arrival rate across three datasets using Vicuna 13B. We measured throughput by limiting each benchmark to 30 minutes and counting completed requests. \alg{} achieves throughput gains over INFERCEPT and vLLM in single-API and multi-API scenarios. In single-API cases, \alg{} outperforms INFERCEPT by up to 46.81\% and vLLM by 71.34\% at higher request rates. The multi-API dataset highlights even greater gains, with \alg{} achieving up to 50.23\% better throughput than INFERCEPT and over 144\% compared to vLLM. However, the improvements over INFERCEPT are less pronounced on the ToolBench dataset due to many requests exceeding 2048 tokens. Both vLLM and INFERCEPT prioritize new requests over ongoing ones, leading to higher throughput for long requests, as seen with ToolBench. In contrast, our system focuses on optimizing end-to-end latency rather than throughput. While this strategy results in gains in latency and throughput on some datasets (such as single-API and multi-API), on the ToolBench dataset, we observe significant improvements in latency accompanied by a slight degradation in throughput.  This trade-off occurs because optimizing latency can sometimes reduce throughput, especially with longer requests.


\begin{figure}[t]
    \centering
    \begin{subfigure}[b]{0.45\linewidth}
        \centering
        \includegraphics[width=\textwidth]{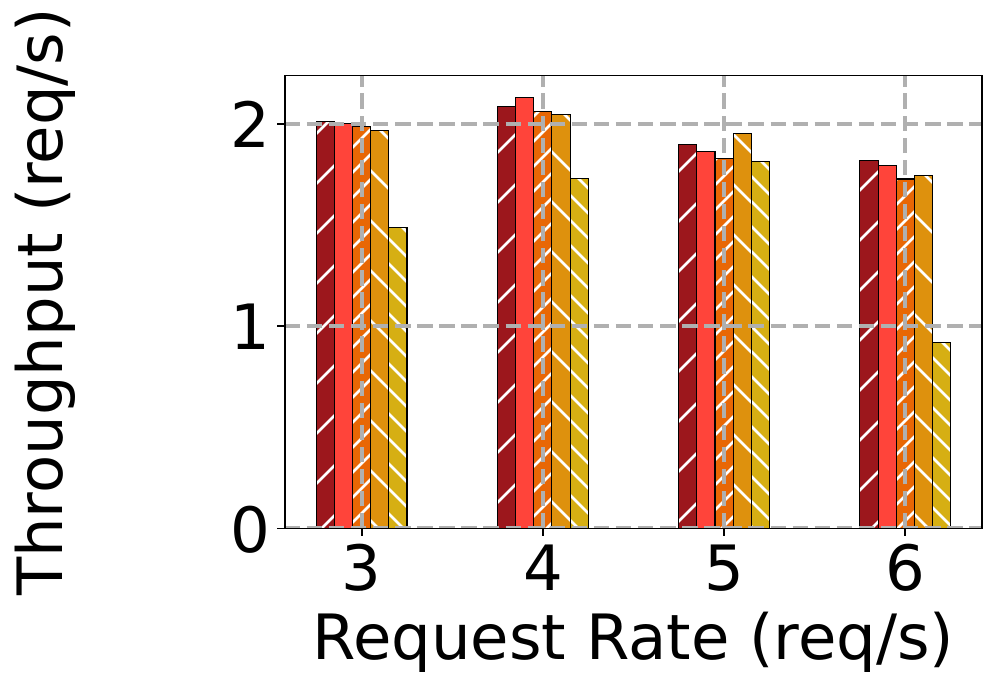}
        \label{fig:err_starvation_throughput}
    \end{subfigure}
    \hfill
    \begin{subfigure}[b]{0.45\linewidth}
        \centering
        \includegraphics[width=\textwidth]{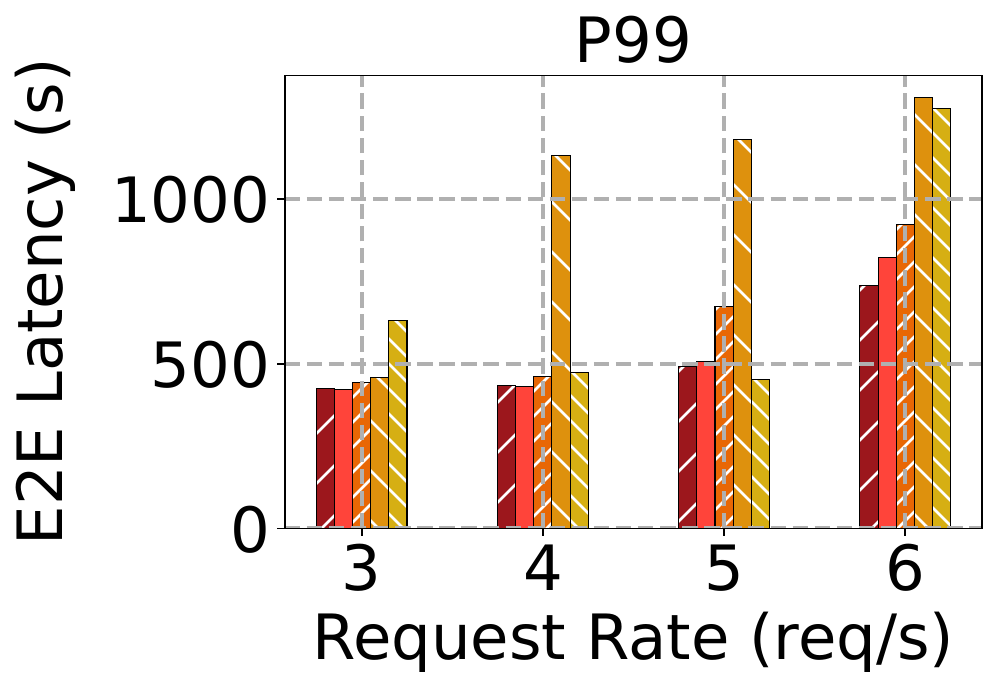}
        \label{fig:err_starvation_latency}
    \end{subfigure}
    \includegraphics[width=\linewidth]{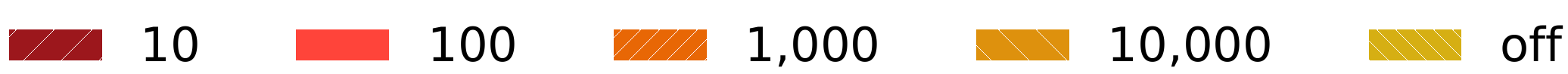}
    \caption{Starvation threshold,  Multi-API dataset with GPT-J 6B.}
    \label{fig:starvation}
\end{figure}

Figure~\ref{fig:starvation} compares the throughput and tail latency of \alg{} under various starvation prevention thresholds. We observe that introducing a starvation prevention threshold reduces tail latency and enhances throughput, with a threshold of 100 providing a good balance between both metrics.

\begin{figure*}[t]
    \centering
    \begin{subfigure}{0.19\textwidth}
        \centering
        \includegraphics[width=\textwidth]{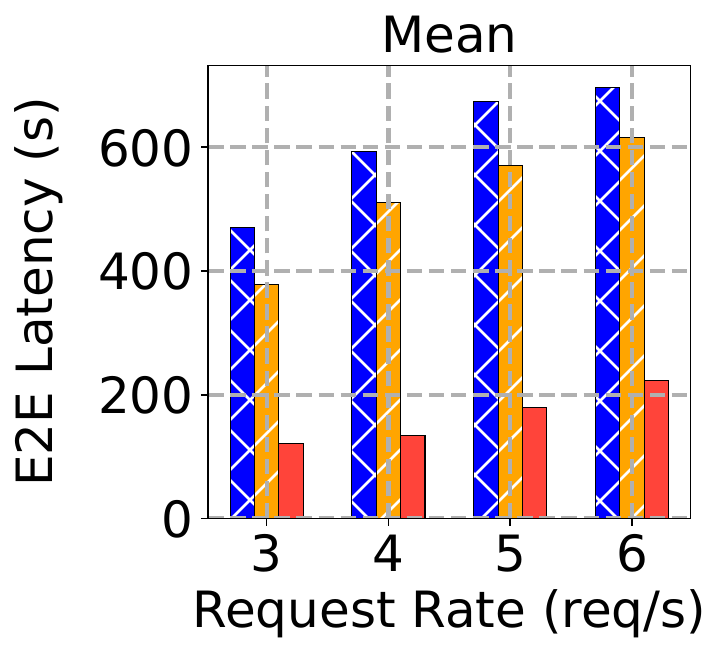}
        \label{fig:subfig1}
    \end{subfigure}
    \begin{subfigure}{0.19\textwidth}
        \centering
        \includegraphics[width=\textwidth]{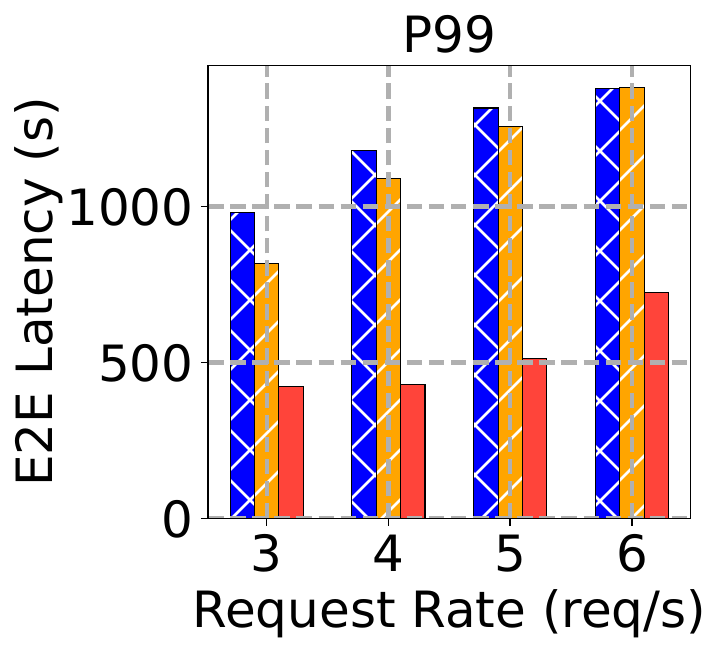}
        \label{fig:subfig3}
    \end{subfigure}
    \begin{subfigure}{0.19\textwidth}
        \centering
        \includegraphics[width=\textwidth]{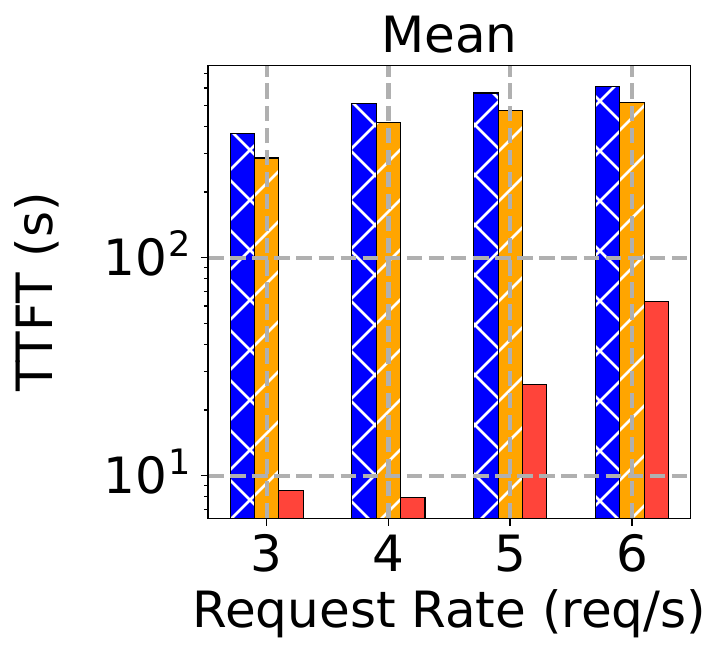}
        \label{fig:subfig4}
    \end{subfigure}
    \begin{subfigure}{0.19\textwidth}
        \centering
        \includegraphics[width=\textwidth]{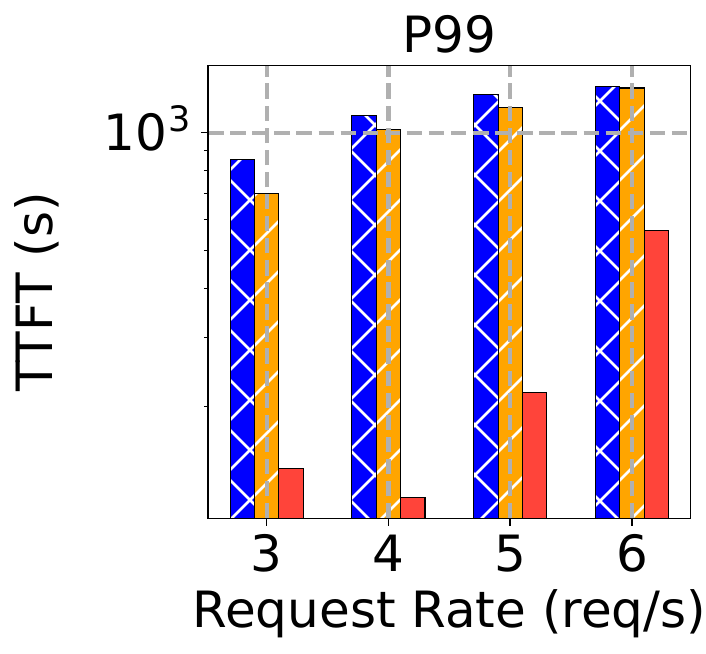}
        \label{fig:subfig6}
    \end{subfigure}
    \begin{subfigure}{0.19\textwidth}
        \centering
        \includegraphics[width=\textwidth]{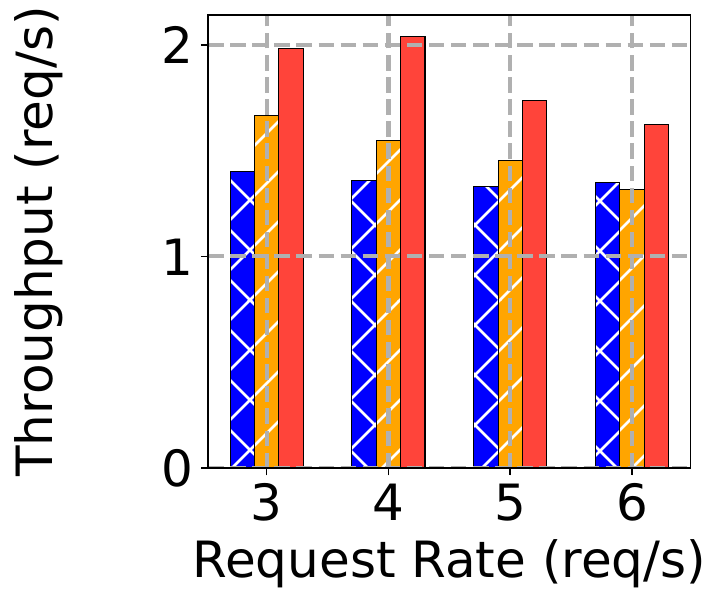}
        \label{fig:subfig7}
    \end{subfigure}

    \includegraphics[width=0.5\textwidth]{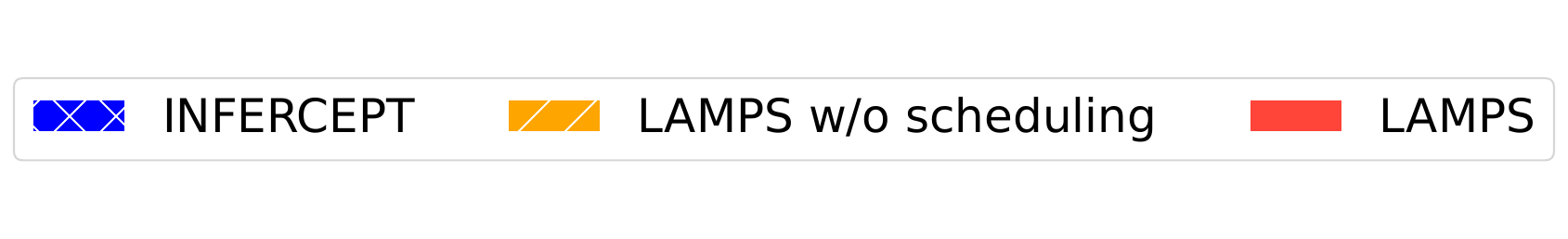}
    \caption{Breakdown of \alg{} components, using Multi-API dataset with Vicuna 13B.}
    \label{fig:perf_breakdown}
\end{figure*}

\subsection{Breakdown of \alg{} Components}
To further understand the benefits of \alg{}, we incrementally added its components to vLLM and compared the results with INFERCEPT. We used the Multi-API dataset because it has the highest latency among the datasets (Figure~\ref{fig:perf_vs_dataset}), Figure~\ref{fig:perf_breakdown} shows throughput, as well as the mean and P99 of end-to-end latency and TTFT.
First, we added the predicted API handling component to vLLM while keeping the scheduling policy as FCFS (referred to as \alg{} w/o scheduling). With this addition, the performance was close to INFERCEPT but slightly worse. The main difference between INFERCEPT and \alg{} w/o scheduling is that we use predicted information to estimate how to handle API calls in advance. In contrast, INFERCEPT dynamically decides how to handle requests during the API call when the request reaches the API.
Next, we integrated our scheduling policy. The main improvements across all metrics came from the scheduling policy. Our scheduling policy effectively reduces head-of-line blocking and optimizes resource utilization. However,  predicting the API handling policy is a necessary preliminary step in implementing our scheduling.

\subsection{Prediction Component}

In this subsection, we evaluate the impact of prediction errors on the performance of \alg{}. 

\begin{figure}[t]
    \centering
    \begin{subfigure}[b]{0.45\linewidth}
        \centering
        \includegraphics[width=\textwidth]{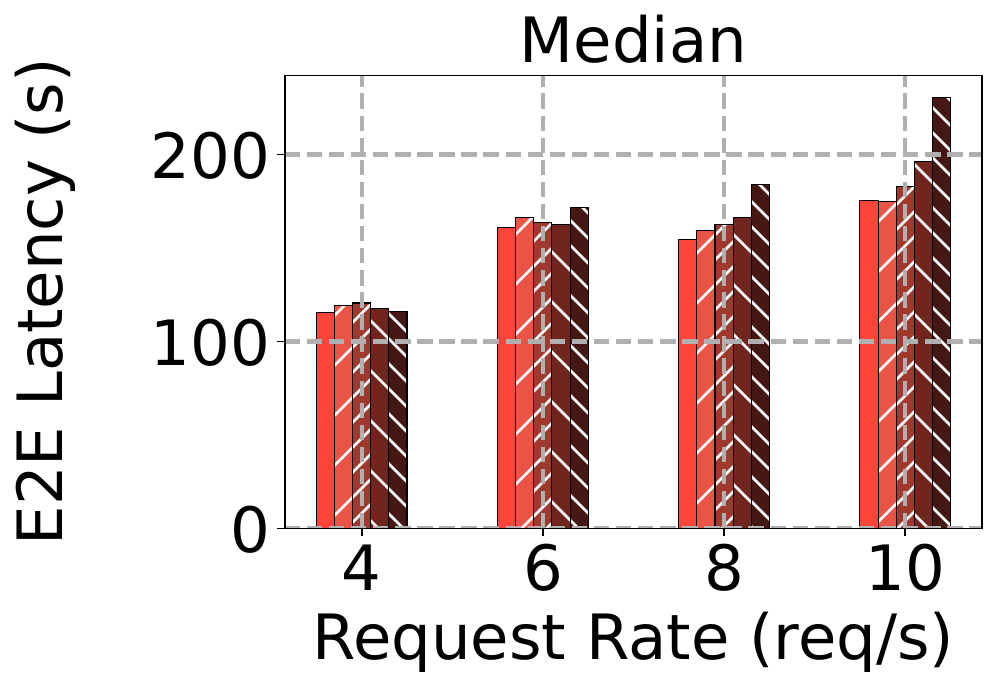}  
        \label{fig:err_injection_median_latency}
    \end{subfigure}
    \hfill
    \begin{subfigure}[b]{0.45\linewidth}
        \centering
        \includegraphics[width=\textwidth]{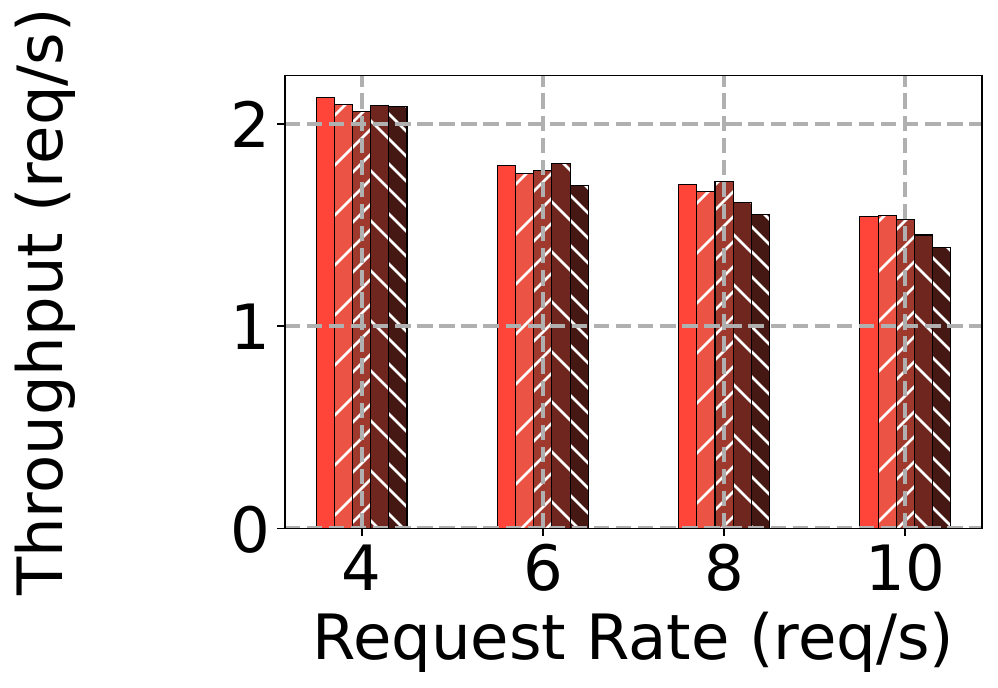}  
        \label{fig:err_injection_throughput}
    \end{subfigure}
    \includegraphics[width=\linewidth]{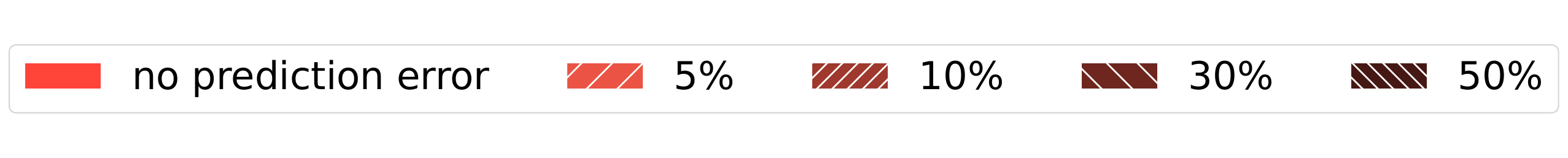}
    \caption{Error injection,  Multi-API dataset with GPT-J 6B.}
    \label{fig:error_injection}
\end{figure}

\textbf{Effect of Mispredictions.} Using the INFERCEPT dataset, we inject controlled Gaussian errors into the predictions for API duration and output length: $\text{error} \sim \mathcal{N}(0, p \times m)$, where
$p$ is the error parameter and $m$ is the measured value. The predicted values are then calculated as: $\text{predicted\_value} = \text{measured\_value} + \text{error}$. By varying the \textit{error\_parameter} parameter, we evaluate how different prediction inaccuracies affect the overall performance of \alg{}. 
Figure ~\ref{fig:error_injection} demonstrates the impact of prediction errors on system performance, focusing on end-to-end latency and throughput. As the prediction error parameter increases (e.g., 5\%, 10\%, 30\%, 50\%), median latency increases, particularly under higher request rates (8-10 req/s). This indicates that inaccurate predictions lead to longer waiting times. Similarly, throughput decreases as error rates increase, especially at higher request rates.
However, we observe that performance degradation in \alg{} occurs only when large prediction errors occur. This suggests that as long as reasonably accurate predictions are maintained, \alg{} can deliver improved performance.

\textbf{Prediction Accuracy and Overhead.}
We evaluated the precision of our response length predictions using the ToolBench dataset by measuring the absolute difference between the predicted and actual word lengths (which is part of the dataset).
We used two accuracy metrics, Acc-5 and Acc-15 that represent the percentage of predictions that differ from the actual length by no more than 5 words and 15 words, respectively.
The results show 68.5\% accuracy for Acc-5 and 78.3\% accuracy for Acc-15, with a Mean Absolute Error (MAE) of 3.06. When focusing on the first 20 bins (responses up to 200 words), the MAE improves to 1.366, indicating higher accuracy for shorter responses.
We used an NVIDIA A100 GPU for inference, achieving an average prediction time of 13.7 ms per input on the ToolBench dataset.
Table~\ref{tb:class_acc} shows Acc-5 and Acc-15 per bin for the first ten bins.

\begin{table}[t]
\centering
\resizebox{\columnwidth}{!}{%
\begin{tabular}{|c|c|c|c|c|c|c|c|c|c|c|c|}
\hline
\textbf{Bin num} & \textbf{0} & \textbf{1} & \textbf{2} & \textbf{3} & \textbf{4} & \textbf{5} & \textbf{6} & \textbf{7} & \textbf{8} & \textbf{9} & \textbf{10} \\ \hline
\textbf{Acc-5} & 0.0 & 0.75 & 0.834 & 0.813 & 0.713 & 0.769 & 0.568 & 0.5 & 0.321 & 0.333 & 0.3 \\ \hline
\textbf{Acc-15} & 0.0 & 0.75 & 0.921 & 0.933 & 0.867 & 0.851 & 0.691 & 0.545 & 0.321 & 0.462 & 0.3 \\ \hline
\end{tabular}%
}
\caption{Bin accuracy for top 10 bins.}
\label{tb:class_acc}
\end{table}

\section{Related Works}
Several studies focus on improving inference throughput through optimized scheduling strategies. Orca~\cite{yu2022orca} introduces iteration-level scheduling, where a new batch is created at the end of each model forward pass. This approach increases GPU utilization and enhances inference throughput. Other research targets efficient GPU memory management. vLLM~\cite{kwon2023efficient} introduces paged attention, treating the KV cache as virtual memory mapped to non-contiguous physical GPU memory, improving GPU memory utilization. Another line of work addresses the imbalance between the prefill and decoding stages. Sarathi~\cite{agrawal2024taming} employs chunked prefill, which divides prompt tokens into smaller chunks merged with decoding requests to form a batch for each iteration. Splitwise~\cite{patel2024splitwise} separates the prefill and decoding stages across different machines to match their distinct computational demands. These techniques for memory optimization are complementary and can be integrated with \alg{}.

Different approaches have been explored to design effective scheduling policies for LLMs.
FastServe~\cite{wu2023fast} builds upon Orca by scheduling each output token individually using a Multi-Level Feedback Queue (MLFQ) to avoid head-of-line blocking. However, this approach leads to frequent preemptions, increasing the cost of managing the key-value (KV) cache memory and offloading to the CPU. Prediction-based scheduling methods have been introduced to address these challenges. Zhen et al.~\citep{zheng2024response} enhanced LLM inference by predicting response lengths with an additional LLM and scheduling requests based on these predictions. While this optimizes latency, it introduces overhead due to the extra LLM used for length prediction.
Other works predict output lengths using lightweight models like DistilBERT and OPT~\cite{jin2023s}. Recently, Trail~\cite{trail} obtained output length predictions directly from the target LLM by feeding the embedding of its internal structures into a lightweight classifier. Using these predictions, they scheduled requests with a prediction-based Shortest Remaining Processing Time variant with limited preemption to manage memory overhead. Importantly, however, all of these previous works focus on requests without API augmentations.

\section{Conclusion and Future Work}

We have introduced \alg{} (\algFULL{}), an LLM inference framework designed explicitly for API-augmented requests. \alg{} optimizes request completion time through a unified scheduling strategy that ranks requests based on their memory consumption over time. By predicting pre-API outputs, \alg{} can estimate the optimal handling strategy during API calls, choosing between preserving, discarding, or swapping memory to minimize memory waste. This predictive approach allows \alg{} to schedule requests with varying output sizes and API interactions effectively. Additionally, our framework incorporates a starvation prevention mechanism for better tail latency. Experimental results demonstrate that \alg{} improves end-to-end latency by 27\%-85\% and reduces TTFT by 4\%-96\% compared to INFERCEPT, with even greater gains over vLLM.

We believe that this work serves as a starting point for API-augmented requests. 
For further improvements, we will focus on enhancing prediction accuracy, particularly for memory-intensive requests, and aim to better handle multi-API requests by accounting for cumulative memory consumption throughout the entire process. Another direction for building upon \alg{} is to manage multiple LLMs, directing requests to the most suitable LLM based on the specific API type and the current load of the LLMs. 
This would be a load-balancing scheduling variation.  Similarly, one might have requests that have to go through a sequence of LLMs and/or servers for API calls, according to some ordering (that differs among requests). This is similar to a jobshop scheduling variation.  

More generally, we suggest that scheduling with API calls appears to open the door to many interesting algorithmic problems.  We are not aware of API calls of the form considered here being studied in the (theoretical) scheduling algorithms literature. More consideration of algorithmic bounds for these types of problems may yield more additional practical strategies for API-augmented requests.


\clearpage

\bibliographystyle{ACM-Reference-Format}
\bibliography{refs}


\begin{thebibliography}{31}


\ifx \showCODEN    \undefined \def \showCODEN     #1{\unskip}     \fi
\ifx \showDOI      \undefined \def \showDOI       #1{#1}\fi
\ifx \showISBNx    \undefined \def \showISBNx     #1{\unskip}     \fi
\ifx \showISBNxiii \undefined \def \showISBNxiii  #1{\unskip}     \fi
\ifx \showISSN     \undefined \def \showISSN      #1{\unskip}     \fi
\ifx \showLCCN     \undefined \def \showLCCN      #1{\unskip}     \fi
\ifx \shownote     \undefined \def \shownote      #1{#1}          \fi
\ifx \showarticletitle \undefined \def \showarticletitle #1{#1}   \fi
\ifx \showURL      \undefined \def \showURL       {\relax}        \fi
\providecommand\bibfield[2]{#2}
\providecommand\bibinfo[2]{#2}
\providecommand\natexlab[1]{#1}
\providecommand\showeprint[2][]{arXiv:#2}

\bibitem[\protect\citeauthoryear{Abhyankar, He, Srivatsa, Zhang, and
  Zhang}{Abhyankar et~al\mbox{.}}{2024}]%
        {abhyankarinfercept}
\bibfield{author}{\bibinfo{person}{Reyna Abhyankar}, \bibinfo{person}{Zijian
  He}, \bibinfo{person}{Vikranth Srivatsa}, \bibinfo{person}{Hao Zhang}, {and}
  \bibinfo{person}{Yiying Zhang}.} \bibinfo{year}{2024}\natexlab{}.
\newblock \showarticletitle{InferCept: Efficient Intercept Support for
  Augmented Large Language Model Inference}. In
  \bibinfo{booktitle}{\emph{Forty-first International Conference on Machine
  Learning}}.
\newblock


\bibitem[\protect\citeauthoryear{Agrawal, Kedia, Panwar, Mohan, Kwatra,
  Gulavani, Tumanov, and Ramjee}{Agrawal et~al\mbox{.}}{2024}]%
        {agrawal2024taming}
\bibfield{author}{\bibinfo{person}{Amey Agrawal}, \bibinfo{person}{Nitin
  Kedia}, \bibinfo{person}{Ashish Panwar}, \bibinfo{person}{Jayashree Mohan},
  \bibinfo{person}{Nipun Kwatra}, \bibinfo{person}{Bhargav~S Gulavani},
  \bibinfo{person}{Alexey Tumanov}, {and} \bibinfo{person}{Ramachandran
  Ramjee}.} \bibinfo{year}{2024}\natexlab{}.
\newblock \showarticletitle{Taming throughput-latency tradeoff in llm inference
  with sarathi-serve}.
\newblock \bibinfo{journal}{\emph{arXiv preprint arXiv:2403.02310}}
  (\bibinfo{year}{2024}).
\newblock


\bibitem[\protect\citeauthoryear{Baeza-Yates, Ribeiro-Neto,
  et~al\mbox{.}}{Baeza-Yates et~al\mbox{.}}{1999}]%
        {baeza1999modern}
\bibfield{author}{\bibinfo{person}{Ricardo Baeza-Yates},
  \bibinfo{person}{Berthier Ribeiro-Neto}, {et~al\mbox{.}}}
  \bibinfo{year}{1999}\natexlab{}.
\newblock \bibinfo{booktitle}{\emph{Modern information retrieval}}.
  Vol.~\bibinfo{volume}{463}.
\newblock \bibinfo{publisher}{ACM press New York}.
\newblock


\bibitem[\protect\citeauthoryear{Betker, Goh, Jing, Brooks, Wang, Li, Ouyang,
  Zhuang, Lee, Guo, et~al\mbox{.}}{Betker et~al\mbox{.}}{2023}]%
        {betker2023improving}
\bibfield{author}{\bibinfo{person}{James Betker}, \bibinfo{person}{Gabriel
  Goh}, \bibinfo{person}{Li Jing}, \bibinfo{person}{Tim Brooks},
  \bibinfo{person}{Jianfeng Wang}, \bibinfo{person}{Linjie Li},
  \bibinfo{person}{Long Ouyang}, \bibinfo{person}{Juntang Zhuang},
  \bibinfo{person}{Joyce Lee}, \bibinfo{person}{Yufei Guo}, {et~al\mbox{.}}}
  \bibinfo{year}{2023}\natexlab{}.
\newblock \showarticletitle{Improving image generation with better captions}.
\newblock \bibinfo{journal}{\emph{Computer Science. https://cdn. openai.
  com/papers/dall-e-3. pdf}} \bibinfo{volume}{2}, \bibinfo{number}{3}
  (\bibinfo{year}{2023}), \bibinfo{pages}{8}.
\newblock


\bibitem[\protect\citeauthoryear{Chase}{Chase}{[n.d.]}]%
        {langchain}
\bibfield{author}{\bibinfo{person}{H.~LangChain Chase}.}
  \bibinfo{year}{[n.d.]}\natexlab{}.
\newblock \bibinfo{title}{LangChain}.
\newblock \bibinfo{howpublished}{\url{ https://github.com/langchain-ai/
  langchain}}.
\newblock


\bibitem[\protect\citeauthoryear{Chen, Chen, Tan, Long, Ga{\v{s}}i{\'c}, and
  Yu}{Chen et~al\mbox{.}}{2019}]%
        {chen2019agentgraph}
\bibfield{author}{\bibinfo{person}{Lu Chen}, \bibinfo{person}{Zhi Chen},
  \bibinfo{person}{Bowen Tan}, \bibinfo{person}{Sishan Long},
  \bibinfo{person}{Milica Ga{\v{s}}i{\'c}}, {and} \bibinfo{person}{Kai Yu}.}
  \bibinfo{year}{2019}\natexlab{}.
\newblock \showarticletitle{AgentGraph: Toward universal dialogue management
  with structured deep reinforcement learning}.
\newblock \bibinfo{journal}{\emph{IEEE/ACM Transactions on Audio, Speech, and
  Language Processing}} \bibinfo{volume}{27}, \bibinfo{number}{9}
  (\bibinfo{year}{2019}), \bibinfo{pages}{1378--1391}.
\newblock


\bibitem[\protect\citeauthoryear{Cheng, Hu, Wang, Du, Li, and Zhang}{Cheng
  et~al\mbox{.}}{2024}]%
        {cheng2024enabling}
\bibfield{author}{\bibinfo{person}{Ke Cheng}, \bibinfo{person}{Wen Hu},
  \bibinfo{person}{Zhi Wang}, \bibinfo{person}{Peng Du},
  \bibinfo{person}{Jianguo Li}, {and} \bibinfo{person}{Sheng Zhang}.}
  \bibinfo{year}{2024}\natexlab{}.
\newblock \showarticletitle{Enabling Efficient Batch Serving for LMaaS via
  Generation Length Prediction}.
\newblock \bibinfo{journal}{\emph{arXiv preprint arXiv:2406.04785}}
  (\bibinfo{year}{2024}).
\newblock


\bibitem[\protect\citeauthoryear{Hao, Liu, Wang, and Hu}{Hao
  et~al\mbox{.}}{2024}]%
        {hao2024toolkengpt}
\bibfield{author}{\bibinfo{person}{Shibo Hao}, \bibinfo{person}{Tianyang Liu},
  \bibinfo{person}{Zhen Wang}, {and} \bibinfo{person}{Zhiting Hu}.}
  \bibinfo{year}{2024}\natexlab{}.
\newblock \showarticletitle{Toolkengpt: Augmenting frozen language models with
  massive tools via tool embeddings}.
\newblock \bibinfo{journal}{\emph{Advances in neural information processing
  systems}}  \bibinfo{volume}{36} (\bibinfo{year}{2024}).
\newblock


\bibitem[\protect\citeauthoryear{Jin, Wu, Brooks, and Wei}{Jin
  et~al\mbox{.}}{2023}]%
        {jin2023s}
\bibfield{author}{\bibinfo{person}{Yunho Jin}, \bibinfo{person}{Chun-Feng Wu},
  \bibinfo{person}{David Brooks}, {and} \bibinfo{person}{Gu-Yeon Wei}.}
  \bibinfo{year}{2023}\natexlab{}.
\newblock \showarticletitle{\$S{\textasciicircum}3\$: Increasing {GPU}
  Utilization during Generative Inference for Higher Throughput}. In
  \bibinfo{booktitle}{\emph{Thirty-seventh Conference on Neural Information
  Processing Systems}}.
\newblock
\urldef\tempurl%
\url{https://openreview.net/forum?id=zUYfbdNl1m}
\showURL{%
\tempurl}


\bibitem[\protect\citeauthoryear{Khattab, Singhvi, Maheshwari, Zhang,
  Santhanam, Haq, Sharma, Joshi, Moazam, Miller, et~al\mbox{.}}{Khattab
  et~al\mbox{.}}{2024}]%
        {khattab2024dspy}
\bibfield{author}{\bibinfo{person}{Omar Khattab}, \bibinfo{person}{Arnav
  Singhvi}, \bibinfo{person}{Paridhi Maheshwari}, \bibinfo{person}{Zhiyuan
  Zhang}, \bibinfo{person}{Keshav Santhanam}, \bibinfo{person}{Saiful Haq},
  \bibinfo{person}{Ashutosh Sharma}, \bibinfo{person}{Thomas~T Joshi},
  \bibinfo{person}{Hanna Moazam}, \bibinfo{person}{Heather Miller},
  {et~al\mbox{.}}} \bibinfo{year}{2024}\natexlab{}.
\newblock \showarticletitle{DSPy: Compiling Declarative Language Model Calls
  into State-of-the-Art Pipelines}. In \bibinfo{booktitle}{\emph{The Twelfth
  International Conference on Learning Representations}}.
\newblock


\bibitem[\protect\citeauthoryear{Kwon, Li, Zhuang, Sheng, Zheng, Yu, Gonzalez,
  Zhang, and Stoica}{Kwon et~al\mbox{.}}{2023}]%
        {kwon2023efficient}
\bibfield{author}{\bibinfo{person}{Woosuk Kwon}, \bibinfo{person}{Zhuohan Li},
  \bibinfo{person}{Siyuan Zhuang}, \bibinfo{person}{Ying Sheng},
  \bibinfo{person}{Lianmin Zheng}, \bibinfo{person}{Cody~Hao Yu},
  \bibinfo{person}{Joseph Gonzalez}, \bibinfo{person}{Hao Zhang}, {and}
  \bibinfo{person}{Ion Stoica}.} \bibinfo{year}{2023}\natexlab{}.
\newblock \showarticletitle{Efficient memory management for large language
  model serving with pagedattention}. In \bibinfo{booktitle}{\emph{Proceedings
  of the 29th Symposium on Operating Systems Principles}}.
  \bibinfo{pages}{611--626}.
\newblock


\bibitem[\protect\citeauthoryear{Mialon, Dess{\`\i}, Lomeli, Nalmpantis,
  Pasunuru, Raileanu, Rozi{\`e}re, Schick, Dwivedi-Yu, Celikyilmaz,
  et~al\mbox{.}}{Mialon et~al\mbox{.}}{2023}]%
        {mialon2023augmented}
\bibfield{author}{\bibinfo{person}{Gr{\'e}goire Mialon},
  \bibinfo{person}{Roberto Dess{\`\i}}, \bibinfo{person}{Maria Lomeli},
  \bibinfo{person}{Christoforos Nalmpantis}, \bibinfo{person}{Ram Pasunuru},
  \bibinfo{person}{Roberta Raileanu}, \bibinfo{person}{Baptiste Rozi{\`e}re},
  \bibinfo{person}{Timo Schick}, \bibinfo{person}{Jane Dwivedi-Yu},
  \bibinfo{person}{Asli Celikyilmaz}, {et~al\mbox{.}}}
  \bibinfo{year}{2023}\natexlab{}.
\newblock \showarticletitle{Augmented language models: a survey}.
\newblock \bibinfo{journal}{\emph{arXiv preprint arXiv:2302.07842}}
  (\bibinfo{year}{2023}).
\newblock


\bibitem[\protect\citeauthoryear{Mitzenmacher}{Mitzenmacher}{2021}]%
        {Mitzenmacher21}
\bibfield{author}{\bibinfo{person}{Michael Mitzenmacher}.}
  \bibinfo{year}{2021}\natexlab{}.
\newblock \showarticletitle{Queues with Small Advice}. In
  \bibinfo{booktitle}{\emph{Proceedings of the 2021 {SIAM} Conference on
  Applied and Computational Discrete Algorithms (ACDA 21)}}.
  \bibinfo{pages}{1--12}.
\newblock


\bibitem[\protect\citeauthoryear{OpenAI}{OpenAI}{2022}]%
        {openai2022chatgpt}
\bibfield{author}{\bibinfo{person}{OpenAI}.} \bibinfo{year}{2022}\natexlab{}.
\newblock \bibinfo{title}{Introducing ChatGPT}.
\newblock \bibinfo{howpublished}{\url{https://openai.com/blog/chatgpt}}.
\newblock


\bibitem[\protect\citeauthoryear{OpenAI.}{OpenAI.}{2023}]%
        {ChatGPTplugins}
\bibfield{author}{\bibinfo{person}{OpenAI.}} \bibinfo{year}{March
  2023}\natexlab{}.
\newblock \bibinfo{title}{ChatGPT plugins}.
\newblock
  \bibinfo{howpublished}{\url{https://openai.com/blog/chatgpt-plugins}}.
\newblock


\bibitem[\protect\citeauthoryear{Patel, Choukse, Zhang, Shah, Goiri, Maleki,
  and Bianchini}{Patel et~al\mbox{.}}{2024}]%
        {patel2024splitwise}
\bibfield{author}{\bibinfo{person}{Pratyush Patel}, \bibinfo{person}{Esha
  Choukse}, \bibinfo{person}{Chaojie Zhang}, \bibinfo{person}{Aashaka Shah},
  \bibinfo{person}{{\'I}{\~n}igo Goiri}, \bibinfo{person}{Saeed Maleki}, {and}
  \bibinfo{person}{Ricardo Bianchini}.} \bibinfo{year}{2024}\natexlab{}.
\newblock \showarticletitle{Splitwise: Efficient generative llm inference using
  phase splitting}. In \bibinfo{booktitle}{\emph{2024 ACM/IEEE 51st Annual
  International Symposium on Computer Architecture (ISCA)}}. IEEE,
  \bibinfo{pages}{118--132}.
\newblock


\bibitem[\protect\citeauthoryear{Patil, Zhang, Wang, and Gonzalez}{Patil
  et~al\mbox{.}}{2023}]%
        {patil2023gorilla}
\bibfield{author}{\bibinfo{person}{Shishir~G Patil}, \bibinfo{person}{Tianjun
  Zhang}, \bibinfo{person}{Xin Wang}, {and} \bibinfo{person}{Joseph~E
  Gonzalez}.} \bibinfo{year}{2023}\natexlab{}.
\newblock \showarticletitle{Gorilla: Large language model connected with
  massive apis}.
\newblock \bibinfo{journal}{\emph{arXiv preprint arXiv:2305.15334}}
  (\bibinfo{year}{2023}).
\newblock


\bibitem[\protect\citeauthoryear{Qin, Liang, Ye, Zhu, Yan, Lu, Lin, Cong, Tang,
  Qian, Zhao, Tian, Xie, Zhou, Gerstein, Li, Liu, and Sun}{Qin
  et~al\mbox{.}}{2023}]%
        {qin2023toolllm}
\bibfield{author}{\bibinfo{person}{Yujia Qin}, \bibinfo{person}{Shihao Liang},
  \bibinfo{person}{Yining Ye}, \bibinfo{person}{Kunlun Zhu},
  \bibinfo{person}{Lan Yan}, \bibinfo{person}{Yaxi Lu}, \bibinfo{person}{Yankai
  Lin}, \bibinfo{person}{Xin Cong}, \bibinfo{person}{Xiangru Tang},
  \bibinfo{person}{Bill Qian}, \bibinfo{person}{Sihan Zhao},
  \bibinfo{person}{Runchu Tian}, \bibinfo{person}{Ruobing Xie},
  \bibinfo{person}{Jie Zhou}, \bibinfo{person}{Mark Gerstein},
  \bibinfo{person}{Dahai Li}, \bibinfo{person}{Zhiyuan Liu}, {and}
  \bibinfo{person}{Maosong Sun}.} \bibinfo{year}{2023}\natexlab{}.
\newblock \bibinfo{title}{ToolLLM: Facilitating Large Language Models to Master
  16000+ Real-world APIs}.
\newblock
\newblock
\showeprint[arxiv]{2307.16789}~[cs.AI]


\bibitem[\protect\citeauthoryear{Qiu, Mao, Patke, Cui, Jha, Wang, Franke,
  Kalbarczyk, Ba{\c{s}}ar, and Iyer}{Qiu et~al\mbox{.}}{2024a}]%
        {qiu2024power}
\bibfield{author}{\bibinfo{person}{Haoran Qiu}, \bibinfo{person}{Weichao Mao},
  \bibinfo{person}{Archit Patke}, \bibinfo{person}{Shengkun Cui},
  \bibinfo{person}{Saurabh Jha}, \bibinfo{person}{Chen Wang},
  \bibinfo{person}{Hubertus Franke}, \bibinfo{person}{Zbigniew Kalbarczyk},
  \bibinfo{person}{Tamer Ba{\c{s}}ar}, {and} \bibinfo{person}{Ravishankar~K
  Iyer}.} \bibinfo{year}{2024}\natexlab{a}.
\newblock \showarticletitle{Power-aware Deep Learning Model Serving with
  $\{$$\mu$-Serve$\}$}. In \bibinfo{booktitle}{\emph{2024 USENIX Annual
  Technical Conference (USENIX ATC 24)}}. \bibinfo{pages}{75--93}.
\newblock


\bibitem[\protect\citeauthoryear{Qiu, Mao, Patke, Cui, Jha, Wang, Franke,
  Kalbarczyk, Ba{\c{s}}ar, and Iyer}{Qiu et~al\mbox{.}}{2024b}]%
        {qiu2024efficient}
\bibfield{author}{\bibinfo{person}{Haoran Qiu}, \bibinfo{person}{Weichao Mao},
  \bibinfo{person}{Archit Patke}, \bibinfo{person}{Shengkun Cui},
  \bibinfo{person}{Saurabh Jha}, \bibinfo{person}{Chen Wang},
  \bibinfo{person}{Hubertus Franke}, \bibinfo{person}{Zbigniew~T Kalbarczyk},
  \bibinfo{person}{Tamer Ba{\c{s}}ar}, {and} \bibinfo{person}{Ravishankar~K
  Iyer}.} \bibinfo{year}{2024}\natexlab{b}.
\newblock \showarticletitle{Efficient interactive LLM serving with proxy
  model-based sequence length prediction}.
\newblock \bibinfo{journal}{\emph{arXiv preprint arXiv:2404.08509}}
  (\bibinfo{year}{2024}).
\newblock


\bibitem[\protect\citeauthoryear{Shahout, Malach, Liu, Jiang, Yu, and
  Mitzenmacher}{Shahout et~al\mbox{.}}{2024}]%
        {trail}
\bibfield{author}{\bibinfo{person}{Rana Shahout}, \bibinfo{person}{Eran
  Malach}, \bibinfo{person}{Chunwei Liu}, \bibinfo{person}{Weifan Jiang},
  \bibinfo{person}{Minlan Yu}, {and} \bibinfo{person}{Michael Mitzenmacher}.}
  \bibinfo{year}{2024}\natexlab{}.
\newblock \showarticletitle{Don’t Stop Me Now: Embedding Based Scheduling for
  LLMs}.
\newblock \bibinfo{journal}{\emph{arXiv preprint arXiv:2410.01035}}
  (\bibinfo{year}{2024}).
\newblock


\bibitem[\protect\citeauthoryear{Shridhar, Yuan, C{\^o}t{\'e}, Bisk, Trischler,
  and Hausknecht}{Shridhar et~al\mbox{.}}{2020}]%
        {shridhar2020alfworld}
\bibfield{author}{\bibinfo{person}{Mohit Shridhar}, \bibinfo{person}{Xingdi
  Yuan}, \bibinfo{person}{Marc-Alexandre C{\^o}t{\'e}},
  \bibinfo{person}{Yonatan Bisk}, \bibinfo{person}{Adam Trischler}, {and}
  \bibinfo{person}{Matthew Hausknecht}.} \bibinfo{year}{2020}\natexlab{}.
\newblock \showarticletitle{Alfworld: Aligning text and embodied environments
  for interactive learning}.
\newblock \bibinfo{journal}{\emph{arXiv preprint arXiv:2010.03768}}
  (\bibinfo{year}{2020}).
\newblock


\bibitem[\protect\citeauthoryear{Stojkovic, Zhang, Goiri, Torrellas, and
  Choukse}{Stojkovic et~al\mbox{.}}{2024}]%
        {stojkovic2024dynamollm}
\bibfield{author}{\bibinfo{person}{Jovan Stojkovic}, \bibinfo{person}{Chaojie
  Zhang}, \bibinfo{person}{{\'I}{\~n}igo Goiri}, \bibinfo{person}{Josep
  Torrellas}, {and} \bibinfo{person}{Esha Choukse}.}
  \bibinfo{year}{2024}\natexlab{}.
\newblock \showarticletitle{Dynamollm: Designing llm inference clusters for
  performance and energy efficiency}.
\newblock \bibinfo{journal}{\emph{arXiv preprint arXiv:2408.00741}}
  (\bibinfo{year}{2024}).
\newblock


\bibitem[\protect\citeauthoryear{Wang, Ma, Feng, Zhang, Yang, Zhang, Chen,
  Tang, Chen, Lin, et~al\mbox{.}}{Wang et~al\mbox{.}}{2024}]%
        {wang2024survey}
\bibfield{author}{\bibinfo{person}{Lei Wang}, \bibinfo{person}{Chen Ma},
  \bibinfo{person}{Xueyang Feng}, \bibinfo{person}{Zeyu Zhang},
  \bibinfo{person}{Hao Yang}, \bibinfo{person}{Jingsen Zhang},
  \bibinfo{person}{Zhiyuan Chen}, \bibinfo{person}{Jiakai Tang},
  \bibinfo{person}{Xu Chen}, \bibinfo{person}{Yankai Lin}, {et~al\mbox{.}}}
  \bibinfo{year}{2024}\natexlab{}.
\newblock \showarticletitle{A survey on large language model based autonomous
  agents}.
\newblock \bibinfo{journal}{\emph{Frontiers of Computer Science}}
  \bibinfo{volume}{18}, \bibinfo{number}{6} (\bibinfo{year}{2024}),
  \bibinfo{pages}{186345}.
\newblock


\bibitem[\protect\citeauthoryear{Wolfram}{Wolfram}{[n.d.]}]%
        {wolfarm}
\bibfield{author}{\bibinfo{person}{Wolfram}.}
  \bibinfo{year}{[n.d.]}\natexlab{}.
\newblock \bibinfo{title}{Chatgpt gets its ‘wolfram superpowers}.
\newblock
  \bibinfo{howpublished}{\url{https://writings.stephenwolfram.com/2023/03/
  chatgpt-gets-its-wolfram-superpowers/}}.
\newblock


\bibitem[\protect\citeauthoryear{Wu, Zhong, Zhang, Huang, Liu, and Jin}{Wu
  et~al\mbox{.}}{2023}]%
        {wu2023fast}
\bibfield{author}{\bibinfo{person}{Bingyang Wu}, \bibinfo{person}{Yinmin
  Zhong}, \bibinfo{person}{Zili Zhang}, \bibinfo{person}{Gang Huang},
  \bibinfo{person}{Xuanzhe Liu}, {and} \bibinfo{person}{Xin Jin}.}
  \bibinfo{year}{2023}\natexlab{}.
\newblock \showarticletitle{Fast distributed inference serving for large
  language models}.
\newblock \bibinfo{journal}{\emph{arXiv preprint arXiv:2305.05920}}
  (\bibinfo{year}{2023}).
\newblock


\bibitem[\protect\citeauthoryear{Yu, Jeong, Kim, Kim, and Chun}{Yu
  et~al\mbox{.}}{2022}]%
        {yu2022orca}
\bibfield{author}{\bibinfo{person}{Gyeong-In Yu}, \bibinfo{person}{Joo~Seong
  Jeong}, \bibinfo{person}{Geon-Woo Kim}, \bibinfo{person}{Soojeong Kim}, {and}
  \bibinfo{person}{Byung-Gon Chun}.} \bibinfo{year}{2022}\natexlab{}.
\newblock \showarticletitle{Orca: A distributed serving system for
  $\{$Transformer-Based$\}$ generative models}. In
  \bibinfo{booktitle}{\emph{16th USENIX Symposium on Operating Systems Design
  and Implementation (OSDI 22)}}. \bibinfo{pages}{521--538}.
\newblock


\bibitem[\protect\citeauthoryear{Zaharia, Khattab, Chen, Davis, Miller, Potts,
  Zou, Carbin, Frankle, Rao, and Ghodsi}{Zaharia et~al\mbox{.}}{2024}]%
        {compound-ai-blog}
\bibfield{author}{\bibinfo{person}{Matei Zaharia}, \bibinfo{person}{Omar
  Khattab}, \bibinfo{person}{Lingjiao Chen}, \bibinfo{person}{Jared~Quincy
  Davis}, \bibinfo{person}{Heather Miller}, \bibinfo{person}{Chris Potts},
  \bibinfo{person}{James Zou}, \bibinfo{person}{Michael Carbin},
  \bibinfo{person}{Jonathan Frankle}, \bibinfo{person}{Naveen Rao}, {and}
  \bibinfo{person}{Ali Ghodsi}.} \bibinfo{year}{2024}\natexlab{}.
\newblock \bibinfo{title}{The Shift from Models to Compound AI Systems}.
\newblock
  \bibinfo{howpublished}{\url{https://bair.berkeley.edu/blog/2024/02/18/compound-ai-systems/}}.
\newblock


\bibitem[\protect\citeauthoryear{Zhang, Roller, Goyal, Artetxe, Chen, Chen,
  Dewan, Diab, Li, Lin, et~al\mbox{.}}{Zhang et~al\mbox{.}}{2022}]%
        {zhang2022opt}
\bibfield{author}{\bibinfo{person}{Susan Zhang}, \bibinfo{person}{Stephen
  Roller}, \bibinfo{person}{Naman Goyal}, \bibinfo{person}{Mikel Artetxe},
  \bibinfo{person}{Moya Chen}, \bibinfo{person}{Shuohui Chen},
  \bibinfo{person}{Christopher Dewan}, \bibinfo{person}{Mona Diab},
  \bibinfo{person}{Xian Li}, \bibinfo{person}{Xi~Victoria Lin},
  {et~al\mbox{.}}} \bibinfo{year}{2022}\natexlab{}.
\newblock \showarticletitle{Opt: Open pre-trained transformer language models}.
\newblock \bibinfo{journal}{\emph{arXiv preprint arXiv:2205.01068}}
  (\bibinfo{year}{2022}).
\newblock


\bibitem[\protect\citeauthoryear{Zheng, Yin, Xie, Huang, Sun, Yu, Cao,
  Kozyrakis, Stoica, Gonzalez, et~al\mbox{.}}{Zheng et~al\mbox{.}}{2023}]%
        {zheng2023efficiently}
\bibfield{author}{\bibinfo{person}{Lianmin Zheng}, \bibinfo{person}{Liangsheng
  Yin}, \bibinfo{person}{Zhiqiang Xie}, \bibinfo{person}{Jeff Huang},
  \bibinfo{person}{Chuyue Sun}, \bibinfo{person}{Cody~Hao Yu},
  \bibinfo{person}{Shiyi Cao}, \bibinfo{person}{Christos Kozyrakis},
  \bibinfo{person}{Ion Stoica}, \bibinfo{person}{Joseph~E Gonzalez},
  {et~al\mbox{.}}} \bibinfo{year}{2023}\natexlab{}.
\newblock \showarticletitle{Efficiently programming large language models using
  sglang}.
\newblock \bibinfo{journal}{\emph{arXiv preprint arXiv:2312.07104}}
  (\bibinfo{year}{2023}).
\newblock


\bibitem[\protect\citeauthoryear{Zheng, Ren, Xue, Luo, Jiang, and You}{Zheng
  et~al\mbox{.}}{2024}]%
        {zheng2024response}
\bibfield{author}{\bibinfo{person}{Zangwei Zheng}, \bibinfo{person}{Xiaozhe
  Ren}, \bibinfo{person}{Fuzhao Xue}, \bibinfo{person}{Yang Luo},
  \bibinfo{person}{Xin Jiang}, {and} \bibinfo{person}{Yang You}.}
  \bibinfo{year}{2024}\natexlab{}.
\newblock \showarticletitle{Response length perception and sequence scheduling:
  An llm-empowered llm inference pipeline}.
\newblock \bibinfo{journal}{\emph{Advances in Neural Information Processing
  Systems}}  \bibinfo{volume}{36} (\bibinfo{year}{2024}).
\newblock


\end{thebibliography}



\end{document}